%% file: acl_latex.tex
\documentclass[11pt]{article}

\usepackage[final]{acl}
\usepackage[table]{xcolor}
\usepackage{booktabs}
\usepackage{tabularx}
\usepackage{pifont}
\usepackage{makecell}
\usepackage{amssymb}
\usepackage{tabularx}      
\usepackage{ragged2e} 
\definecolor{lightgreen}{RGB}{230,245,234}
\definecolor{lightred}{RGB}{252,235,235}
\definecolor{headergray}{RGB}{245,245,245}
\definecolor{clagblue}{RGB}{35,85,155}
\usepackage{times}
\usepackage{latexsym}
\usepackage{tcolorbox}
\usepackage{mathtools}

\usepackage{subcaption}   
\usepackage[T1]{fontenc}

\usepackage[utf8]{inputenc}

\usepackage{microtype}
\usepackage{tcolorbox}
\tcbuselibrary{skins, breakable} 
\usepackage{siunitx}

\usepackage{booktabs}
\usepackage{multirow}
\usepackage{makecell}
\usepackage{graphicx}   
\usepackage{tabularx}   
\usepackage{array}      
\tcbset{promptbox/.style={
    colback=white,      
    colframe=black,     
    arc=3mm,            
    boxrule=1pt,        
    fonttitle=\bfseries, 
    fontupper=\ttfamily\footnotesize, 
    left=5pt, right=5pt, top=5pt, bottom=5pt, 
    breakable           
}}
\usepackage{inconsolata}
\usepackage{graphicx}
\usepackage{amsmath}
\usepackage[ruled, linesnumbered, noend]{algorithm2e}
\title{CLAG: Adaptive Memory Organization via Agent-Driven\\ Clustering for Small Language Model Agents}

\author{
  \vspace{0.5em} 
  \textbf{Taeyun Roh}\textsuperscript{1} \quad
  \textbf{Wonjune Jang}\textsuperscript{2} \quad
  \textbf{Junha Jung}\textsuperscript{1,3} \quad
  \textbf{Jaewoo Kang}\textsuperscript{1,3,$\dagger$} \\[0.6em] 
  \textsuperscript{1}Korea University \quad
  \textsuperscript{2}Myongji University \quad
  \textsuperscript{3}AIGEN Sciences \\[0.4em] 
  \texttt{\{nrbsld, goodjungjun, kangj\}@korea.ac.kr} \\
  \texttt{dnjswnswkd03@mju.ac.kr} 
}
\begin{document}
\maketitle
\begin{abstract}
Large language model agents heavily rely on external memory to support knowledge reuse and complex reasoning tasks. Yet most memory systems store experiences in a single global retrieval pool which can gradually dilute or corrupt stored knowledge. This problem is especially pronounced for small language models (SLMs), which are highly vulnerable to irrelevant context. We introduce \textbf{CLAG}, a \textbf{CL}ustering-based \textbf{AG}entic memory framework where an SLM agent actively organizes memory by clustering. CLAG employs an SLM-driven router to assign incoming memories to semantically coherent clusters and autonomously generates cluster-specific profiles—including topic summaries and descriptive tags—to establish each cluster as a self-contained functional unit. By performing localized evolution within these structured neighborhoods, CLAG effectively reduces cross-topic interference and enhances internal memory density. During retrieval, the framework utilizes a two-stage process that first filters relevant clusters via their profiles, thereby excluding distractors and reducing the search space. Experiments on multiple QA datasets with three SLM backbones show that CLAG consistently improves answer quality and robustness over prior memory systems for agents, remaining lightweight and efficient.
\end{abstract}
\def\thefootnote{$\dagger$}\footnotetext{Corresponding author.}\def\thefootnote{\arabic{footnote}}
\newcommand\blfootnote[1]{%
  \begingroup
  \renewcommand\thefootnote{}\footnote{#1}%
  \addtocounter{footnote}{-1}%
  \endgroup
}
\blfootnote{Our code is available at \url{https://github.com/dmis-lab/CLAG}}
\begin{figure*}[t]
 \centering
 \includegraphics[width=0.9\textwidth]{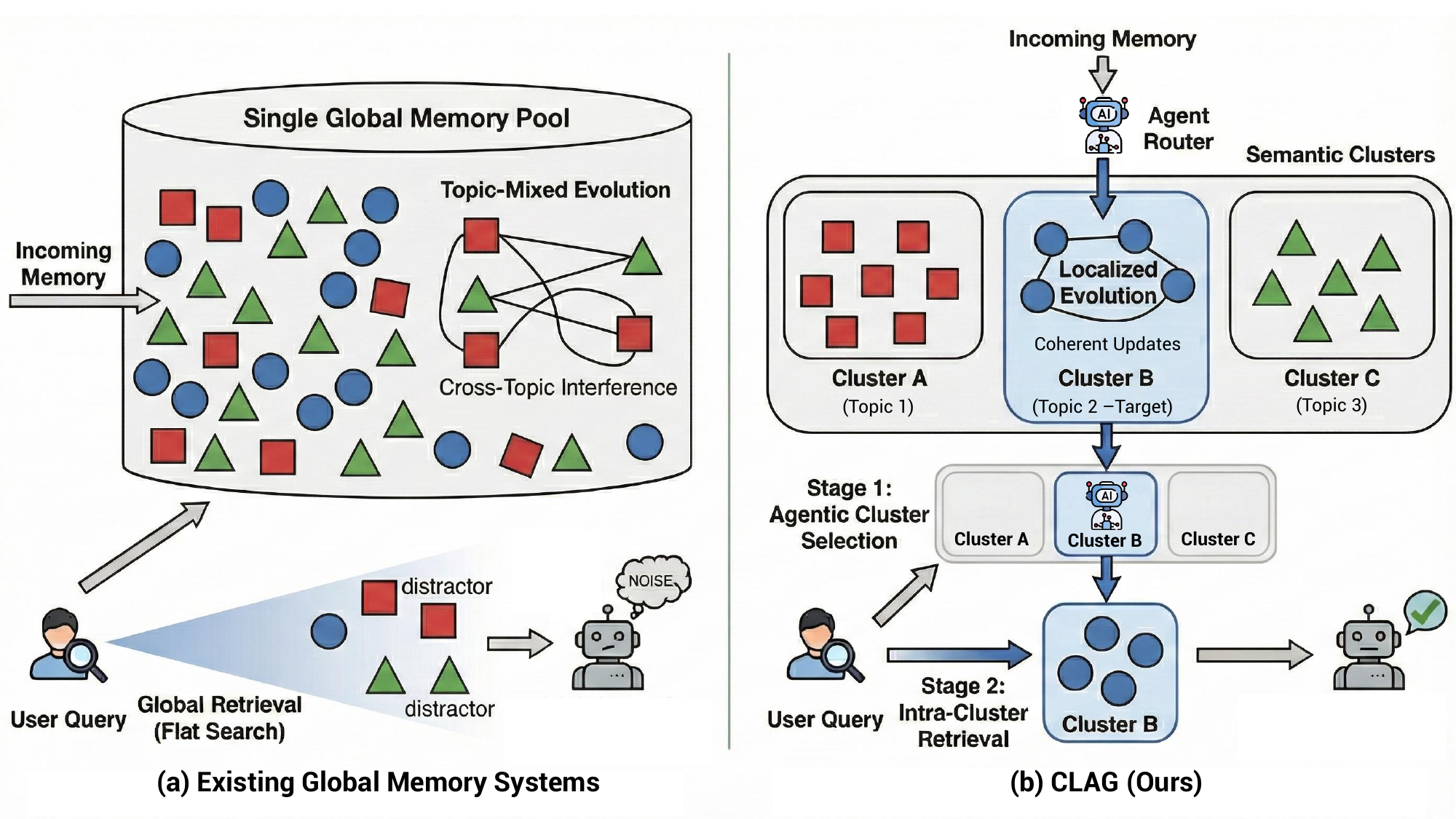}
\caption{Conceptual comparison between existing global memory systems and CLAG. \textbf{(Left)} Traditional approaches manage memories in a single global pool, where topic-mixed updates and retrieval lead to high interference and noise accumulation. \textbf{(Right)} CLAG employs agent-driven clustering to organize memories into semantically coherent neighborhoods. By confining evolution and retrieval to these local clusters, our framework significantly reduces cross-topic interference and enhances memory stability.}

\label{fig:motivation_figure}
\end{figure*}

\section{Introduction}

AI agents are becoming the primary paradigm for deploying Large Language Models (LLMs) in real-world, long-horizon tasks~\cite{xi2023rise,zhou2023webarena,li2023camel,wang2024survey}. To succeed in these settings, LLMs rely on internal memory mechanisms to support knowledge reuse, multi-session coherence, and complex reasoning tasks ~\cite{wang2024survey,xi2025rise}. Conventionally, basic memory systems for AI agents adhere to a standard Retrieval Augmented Generation (RAG) \cite{lewis2020retrieval} paradigm. In this setup, memory functions essentially as a static repository designed to overcome fixed context window limitations. Agents sequentially record raw interaction logs, covering environmental observations, executed actions, tool output, and feedback loops during task execution~\cite{shinn2023reflexion,yao2023reactsynergizingreasoningacting}. When faced with a new state, the agent simply retrieves relevant historical snippets based on semantic similarity to ground its current input, treating memory merely as a fixed-access database of past episodes. Consequently, this static approach prevents the agent from learning from subsequent feedback or refining outdated information based on new experiences~\cite{zhong2023memorybankenhancinglargelanguage,wang2023voyageropenendedembodiedagent}. Moving beyond static storage, frameworks demonstrating \textit{agentic memory }~\cite{xu2025amemagenticmemoryllm,kang2025memory,yan2025general} integrate active agents into memory management to enable continuous self-evolution. However, as illustrated in Figure~\ref{fig:motivation_figure}, these systems typically operate within a \textbf{single global memory pool}, where both evolution and retrieval processes are conducted across the entire unstructured storage. Consequently, as a memory buffer grows, retrieval becomes vulnerable to two coupled issues: (i) the search space expands, increasing the probability of retrieving semantically plausible but task-irrelevant memories, and (ii) memory evolution mechanisms are exposed to topic-mixed neighborhoods, which can misguide updates and gradually degrade the memory store.  These problems are especially salient for small language models (SLMs), which are highly vulnerable to irrelevant context~\cite{mallen2023not,shi2023large,yoran2023making,lu2024small}.

To address these limitations, we propose \textbf{CLAG}, a \textbf{CL}ustering-based \textbf{AG}entic memory framework that imposes lightweight structure on long-horizon memory while preserving the self-evolving nature of modern agentic systems. The core idea is to treat clustering as an \textit{agent-controlled operation} rather than a static offline preprocessing step. Upon memory write, the agent assigns each new memory to a semantically coherent cluster through SLM-agent guided routing and maintains cluster-specific profiles. This process establishes topic-consistent neighborhoods where memory evolution is conducted locally. This design aligns with cognitive principles suggesting that new information should refine relevant schemas without perturbing unrelated memory structures~\cite{tse2007schemas,kumaran2016learning,gilboa2017neurobiology}. By confining evolution to these semantic neighborhoods, CLAG prevents noisy updates from propagating across the global store, allowing each cluster to act as a self-organizing unit that continuously enhances its internal coherence through local interactions.

At inference time, CLAG adopts a \textbf{two-stage retrieval strategy}. Given a query, the agent first selects a small set of relevant clusters using centroid-based filtering and an agentic cluster-selection module, and then retrieves fine-grained memories only within the selected clusters. This hierarchical design reduces the retrieval scope and excludes semantically plausible but task-irrelevant memories that are more likely to appear under global retrieval. Across multiple QA benchmarks and three SLM backbones, we find that this structured control loop yields consistent improvements in answer quality and robustness compared to prior agent memory baselines, while remaining lightweight in both computation and storage. These gains align with our analysis: clustering produces cleaner semantic neighborhoods for evolution and retrieval, and the two-stage retrieval pipeline mitigates the distractor exposure inherent in global retrieval.

Our contributions are threefold:
\begin{itemize}
    \item We introduce an \textbf{agent-driven clustering} mechanism that organizes memories into semantically coherent groups via SLM-agent-based routing, ensuring robust long-horizon organization.
    \item We propose \textbf{localized memory evolution} that updates and consolidates memories within topic-consistent neighborhoods to stabilize long-term memory quality and mitigate cross-topic interference.
    \item We develop a \textbf{two-stage} cluster-aware retrieval scheme that improves robustness for limited-capacity models by reducing the search space, suppressing distractors, and lowering retrieval noise.
\end{itemize}

\section{Related Work}
\subsection{Retrieval Augmented Generation}
Retrieval Augmented Generation (RAG)~\cite{lewis2020retrieval} extends the context window of LLMs by coupling a parametric model with a non-parametric external knowledge base. In agentic settings, the knowledge base is replaced by a memory serving as a repository for interaction histories with the environment. Typically, agents rely on retrieval over these accumulated records to inform current decision-making~\cite{park2023generative,zhang2025survey,yao2023reactsynergizingreasoningacting}.

While variants such as hierarchical indexing~\cite{sarthi2024raptorrecursiveabstractiveprocessing,rezazadeh2024isolated,edge2025localglobalgraphrag,wang2025scmenhancinglargelanguage,xi2023risepotentiallargelanguage} or query rewriting ~\cite{ma2023query,wang2023query2doc,wang2025speculativeragenhancingretrieval} improve coverage, most standard RAG systems operate on a \emph{read-only} basis over a static index. They typically process queries as independent, stateless events, lacking the mechanism to dynamically consolidate or restructure memory based on the agent's continuous experience~\cite{packer2024memgptllmsoperatingsystems}.
\subsection{Memory system for LLM agents}
As LLMs are increasingly deployed as AI agents in long-horizon settings, retrieval has been repurposed into {memory for LLM agents}: storing interaction traces (observations, actions, tool outputs, feedback) and retrieving relevant past snippets to guide current decisions~\cite{yao2023reactsynergizingreasoningacting,zhang2025survey}. Early agent memories largely followed a {static} RAG-style design, treating memory as an append-only repository queried by global similarity search, which limits learning from feedback and consolidation over time. Recent {agentic} and {evolving} memory systems (e.g., A-mem~\cite{xu2025amemagenticmemoryllm}, MemoryOS~\cite{kang2025memory}, GAM~\cite{yan2025general}) introduce maintenance operations such as reflection, compression, and rewriting, but typically still rely on a single {global} memory pool. In contrast, we reduce reliance on global search by \emph{agentically} clustering memories online and routing new entries to an appropriate cluster, conditioning retrieval and memory operations on cluster-level structure.



\begin{figure*}[t]
    \centering
    \includegraphics[width=\textwidth]{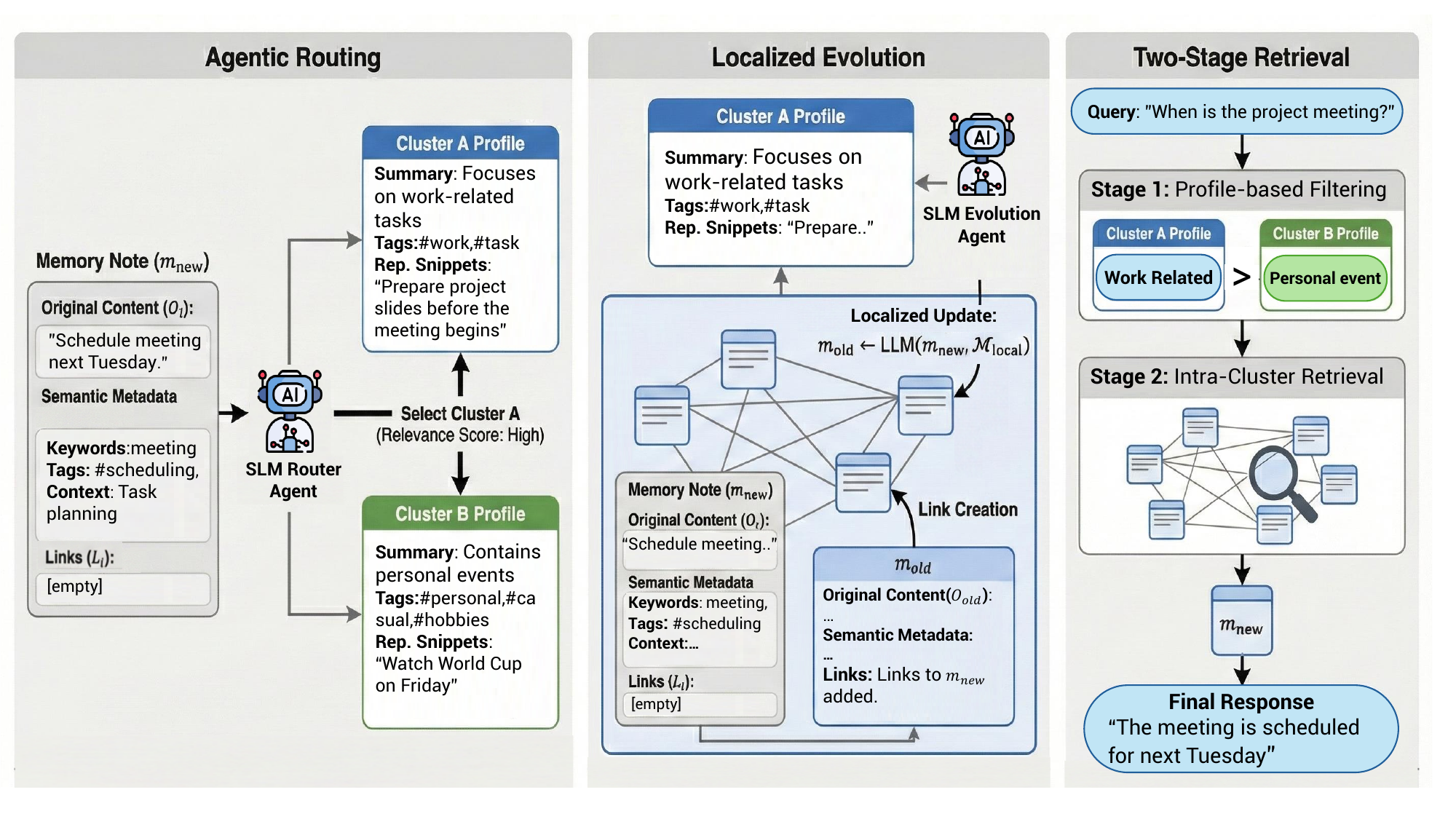}
    \caption{\textbf{Overview of the proposed CLAG framework.} \textbf{Left: Agentic Routing.} An SLM router assigns each incoming memory note $m_{\text{new}}$ to the most relevant cluster using semantic metadata, and updates the corresponding cluster profile $\mathcal{P}$. \textbf{Middle: Localized Evolution.} An evolution agent performs consolidation (e.g., linking, rewriting, strengthening) \emph{within the routed cluster} to maintain topic-consistent neighborhoods and reduce cross-topic interference. \textbf{Right: Two-Stage Retrieval.} Given a query, CLAG first filters clusters using profile-based selection (Stage~1), then retrieves fine-grained memories only inside the selected clusters (Stage~2), reducing the effective search space and suppressing retrieval noise.}
    \label{fig:main_figure}
\end{figure*}

\section{CLAG}
\label{sec:clag}

As illustrated in Figure~\ref{fig:main_figure}, the CLAG framework comprises three core components designed to structure and manage long-horizon memory: (1) \textbf{Agentic Routing}, which assigns incoming memories to semantically coherent clusters; (2) \textbf{Localized Evolution}, which consolidates information within specific clusters to maintain topic consistency; and (3) \textbf{Cluster-Aware Two-Stage Retrieval}, which filters irrelevant clusters to suppress noise during inference. All agentic decisions in CLAG (routing, evolution, and cluster selection) are produced by the same backbone SLM, invoked multiple times with role-specific prompts (router/evolver/selector), rather than separate learned models.

\subsection{Memory Structuring}
\label{sec:memory_structuring}

Inspired by recent agentic memory frameworks such as A-mem \cite{xu2025amemagenticmemoryllm}, 
we represent an agent's long-term memory as a collection of structured {memory notes} that
encode both the raw interaction trace and SLM-agent generated semantic annotations.
Formally, let $\mathcal{M} = \{m_1, m_2, \dots, m_N\}$ denote the set of all memory notes.
Each note $m_i$ is represented as
\begin{equation}
    m_i = \{O_i, C_i, t_i, K_i, G_i, X_i, L_i\}
    \label{eq:note_def}
\end{equation}
where $O_i$ is the original interaction content, $C_i$ is the cluster that the memory belongs to, $t_i$ is the timestamp, $K_i$ is a set of SLM-agent generated keywords that summarize salient concepts, $G_i$ is a set of SLM-agent generated tags for additional information, $X_i$ is an SLM-generated contextual description that captures higher-level semantics, and $L_i$ stores links to other notes that are semantically related.
\subsection{Memory Routing}
\label{sec:memory_routing}
We employ an SLM agent to route each incoming memory to the most semantically relevant cluster. The procedure detailed in Algorithm \ref{alg:memory_routing} operates in two phases based on the number of processed memories $N_{processed}$. During the initial "cold-start" phase (when $N_{processed} < n$), memories are buffered until sufficient data is collected to establish the initial cluster structure $\mathcal{C}$ via the \text{InitializeClusters} routine. Once initialized, subsequent memories undergo a routing process. First, a coarse filtering step identifies the top-$k$ candidate clusters ($\mathcal{C}_{topK}$) closest to the new memory based on vector distance. Given the candidate clusters, the final selection is delegated to the SLM agent, which reviews the semantic profiles of these candidates to select the most appropriate cluster $C_{routed}$. Finally, a cosine similarity check against a threshold $\tau$ determines if the memory is routed to $C_{routed}$ or if it represents novel information requiring the creation of a new cluster $C_{new}$.  

To ensure scalability and prevent semantic drift, we employ an {adaptive clustering} mechanism (Algorithm~\ref{alg:split_cluster}). When a cluster $C_{target}$ exceeds a size threshold $\tau_{split}$, it is dynamically bisected into two sub-clusters via K-Means. This on-the-fly refinement mitigates over-saturation, keeping the memory structure adaptive to the agent’s evolving experiences.

\begin{algorithm}[t]
\SetKw{KwInput}{Input:}
\SetKw{KwParam}{Parameters:}
\SetKwComment{Comment}{// }{}

\caption{Agentic Routing for New Memory}
\label{alg:memory_routing}

\noindent \KwInput{} New memory data $m_{new}$, Current set of clusters $\mathcal{C}$, Count of processed memories $N_{processed}$ \\
\noindent \KwParam{} Initialization size $n$, Routing top-k $k$, Similarity threshold $\tau$, SLM client $SLM$

\BlankLine 

\uIf{$N_{processed} < n$}{
    Add $m_{new}$ to initialization buffer $\mathcal{B}$\;
    \If{$|\mathcal{B}| == n$}{
        $\mathcal{C} \leftarrow \text{InitializeClusters}(\mathcal{B})$\;
    }
}
\Else{
    Identify set $\mathcal{C}_{topK}$ containing $k$ closest clusters\;
    $C_{routed} \leftarrow SLM.\text{SelectCluster}(m_{new}, \mathcal{C}_{topK})$\;
    $sim = \text{CosineSimilarity}(\mathbf{m}_{new}, {C}_{routed})$\;
    \eIf{$sim < \tau$}{
        Create new cluster $C_{new}$ initialized with $m_{new}$\;
        $\mathcal{C} \leftarrow \mathcal{C} \cup \{C_{new}\}$\;
    }{
        Add $m_{new}$ to $C_{routed}$\;
    }
}
$N_{processed} \leftarrow N_{processed} + 1$\;

\end{algorithm}
\subsection{Localized Evolution within Clusters}
\label{sec:localized_evolution}

A distinct advantage of CLAG is that both link generation and memory evolution operate strictly within the assigned cluster $C_{\text{routed}}$. This localized strategy not only ensures that connections are formed within a coherent semantic context but also reduces computational overhead by precluding global pairwise comparisons.

Upon the assignment of a newly routed memory $m_{\text{new}}$ to $C_{\text{routed}}$, CLAG identifies the top-$k$ most relevant neighbors within the cluster. We denote this subset of semantically aligned memories as $\mathcal{M}_{\text{local}}$, defined via cosine similarity:
\begin{equation}
    \mathcal{M}_{\text{local}} = \{ m_j \in C_{\text{routed}} \mid \text{rank}(s_{\text{new},j}) \le k \}
\end{equation}
where $s_{\text{new},j}$ is the cosine similarity between $\mathbf{m}_{\text{new}}$ and $\mathbf{m}_{j}$.

\paragraph{Intra-Cluster Linking}
CLAG establishes explicit connections between the new memory and the retrieved candidates. Since $\mathcal{M}_{\text{local}}$ is constructed via cluster routing and similarity ranking, it inherently contains memories that are semantically proximate yet often difficult to distinguish by vector similarity alone. To address this, SLM agent analyzes the fine-grained relationships (e.g., causality, temporal sequence) between $m_{\text{new}}$ and $\mathcal{M}_{\text{local}}$ to generate a set of links $L_{\text{new}}$:
\begin{equation}
    L_{\text{new}} \leftarrow \text{SLM}(m_{\text{new}} \parallel \mathcal{M}_{\text{local}})
\end{equation}

\begin{table*}[t]
\centering
\renewcommand{\arraystretch}{0.89} 
\setlength{\tabcolsep}{3.5pt}

\setlength{\aboverulesep}{0.3ex} 
\setlength{\belowrulesep}{0.3ex}

\resizebox{0.88\textwidth}{!}{%
\begin{tabular}{ll cc cc cc}
    \toprule
    \multirow{2}{*}{\textbf{Model}} & \multirow{2}{*}{\textbf{Method}} &
    \multicolumn{2}{c}{\textbf{LoCoMo}} &
    \multicolumn{2}{c}{\textbf{HotpotQA}} &
    \multicolumn{2}{c}{\textbf{BioASQ}} \\
    \cmidrule(lr){3-4}\cmidrule(lr){5-6}\cmidrule(lr){7-8}
    & & \textbf{F1} & \textbf{BLEU-1}
      & \textbf{F1} & \textbf{BLEU-1}
      & \textbf{F1} & \textbf{BLEU-1} \\
    \midrule
    
    \multirow{5}{*}{\rotatebox[origin=c]{90}{\textbf{Qwen3-0.6B}}}
    & RAG        & 12.9 & 10.39 & 11.75 & \underline{11.17} & 2.4 & 1.71 \\
    & A-mem      & 14.29 & 11.8 & \underline{12.04}  & 10.65  & \underline{3.61} & 2.83 \\
    & GAM & \underline{16.05} & \underline{13.24} & 7.81 & 6.69 & 3.40 & \underline{3.37} \\
    & MemoryOS   & 4.30  & 3.24  & 9.02  & 7.34  & 3.12 & 1.29 \\
    \cmidrule(lr){2-8}
    & \textbf{CLAG (Ours)} 
    & \textbf{20.99} & \textbf{17.88}
    & \textbf{15.50} & \textbf{14.33}
    & \textbf{22.01} & \textbf{17.23} \\
    \midrule

    \multirow{5}{*}{\rotatebox[origin=c]{90}{\makecell{\textbf{Llama3.2-1B}\\\textbf{Instruct}}}}
    & RAG        & 18.04 & 15.29 & 9.46  & 8.19  & 5.12 & 3.42 \\
    & A-mem      & 14.80 & 12.20 & 11.19 & 10.01 & 5.38 & \underline{4.95} \\
    & GAM & \textbf{22.63} & \textbf{19.85} & \underline{13.85} & \textbf{12.84} & \underline{6.52} & 4.71 \\
    & MemoryOS   & 9.67  & 6.03  & 6.43  & 4.76  & 3.55 & 2.52 \\
    \cmidrule(lr){2-8}
    & \textbf{CLAG (Ours)} 
    & \underline{21.05} & \underline{18.16}
    & \textbf{14.20} & \underline{12.30}
    & \textbf{10.16} & \textbf{8.08} \\
    \midrule

    \multirow{5}{*}{\rotatebox[origin=c]{90}{\makecell{\textbf{DeepSeek-R1}\\\textbf{Distill-Qwen}\\\textbf{1.5B}}}}
    & RAG        & 11.54 & 9.13  & 5.54  & 4.41  & 2.81 & 2.12 \\
    & A-mem      & \underline{12.50} & \underline{9.83} & 6.06 & \underline{4.81}  & \underline{4.45} & \underline{3.07} \\
    & GAM & 12.34 & 9.27 & \underline{6.45} & 4.55 & 2.78 & 2.19 \\
    & MemoryOS   & 6.41  & 4.76  & 4.51  & 2.84  & 2.93 & 1.63 \\
    \cmidrule(lr){2-8}
    & \textbf{CLAG (Ours)} 
    & \textbf{15.46} & \textbf{12.6}
    & \textbf{9.12} & \textbf{6.63}
    & \textbf{6.18} & \textbf{4.42} \\
    
    \bottomrule
\end{tabular}%
}
\caption{Overall results on LoCoMo, HotpotQA, and BioASQ across three backbone models. Best results are marked in \textbf{bold}, 2nd-best results are marked in \underline{underline}.}
\label{tab:overall_results}
\end{table*}

\begin{table}[h]
    \centering
    \small
    \begin{tabular}{l c c}
    \toprule
    \textbf{Method} & \textbf{Retrieval (ms)} & \textbf{End-to-End (ms)} \\
    \midrule
    RAG & 17.80 & 289.60 \\
    A-mem & 18.54 & 408.68 \\
    GAM & 8303.41 & 17934.32   \\ 
    MemoryOS & 220.79 & 968.75 \\
    \midrule
    \textbf{CLAG (Ours)} & 142.43 & 514.14 \\
    \bottomrule
    \end{tabular}
    \caption{Latency comparison on Qwen3-0.6B backbone. All measurements were conducted on a single NVIDIA A100 GPU. Values represent the average time in milliseconds (ms).}
    \label{tab:latency}
\end{table}

\paragraph{Localized Evolution}
Following link generation, the system creates a feedback loop to refine existing knowledge. For each neighbor $m_j$ in $\mathcal{M}_{\text{local}}$, the system determines if its attributes require an update to reflect the new information:
\begin{equation}
    m^*_j \leftarrow \text{SLM}( m_{\text{new}} \parallel \mathcal{M}_{\text{local}} \setminus \{m_j\} \parallel m_j )
\end{equation}
The evolved memory $m^*_j$ then replaces $m_j$ within the cluster. Concurrently, the cluster profile is updated to maintain alignment with the refined semantic state of the cluster. This localized evolution mimics human cognitive processes, where new experiences primarily reshape our understanding of related concepts (i.e., the cluster) without disrupting unrelated knowledge domains~\cite{tse2007schemas,kumaran2016learning,gilboa2017neurobiology}. Consequently, each cluster functions as a self-refining module that iteratively enhances its semantic density and internal coherence through localized updates.

\subsection{Cluster-Aware Two-Stage Retrieval}
\label{sec:two_stage_retrieval}

At inference time, CLAG replaces flat global retrieval with a {cluster-aware two-stage retrieval pipeline}.  
Given a user query $q$, the system initially identifies a candidate subset of relevant clusters and subsequently performs retrieval only within those clusters.

\paragraph{Stage 1: Agentic Cluster Selection.}
The query and its tags are concatenated and embedded into a vector $\mathbf{e}_q$. Based on the distance to cluster centroids $\{\mathbf{c}_k\}$, a candidate set $\mathcal{C}_{\text{cand}}$ of the $K$ closest clusters is identified. Instead of a deterministic top-$K$ selection, the final decision is delegated to the SLM agent. By evaluating cluster profiles against the query, agent returns a variable-size subset $\mathcal{C}_{\text{selected}} \subseteq \mathcal{C}_{\text{cand}}$, constraining the search space and suppressing irrelevant noise, which mitigates vulnerability to distractors.

\paragraph{Stage 2: Intra-Cluster Retrieval.}
Given the selected cluster set $\mathcal{C}_{\text{selected}}$, CLAG performs retrieval only within the union of their members
$\mathcal{M}_{\text{cand}} = \bigcup_{C \in \mathcal{C}_{\text{selected}}} C$. By restricting retrieval to cluster-local neighborhoods, CLAG reduces the effective search space and suppresses semantically plausible but task-irrelevant memories that commonly arise under global similarity search. This hierarchical procedure is particularly beneficial for SLM agents: the agentic cluster-selection step provides a high-level semantic filter, improving robustness to retrieval noise.
\begin{table*}[t]
\centering
\renewcommand{\arraystretch}{0.95}
\setlength{\tabcolsep}{2.5pt}

\resizebox{0.95\textwidth}{!}{%
\begin{tabular}{ll cc cc cc cc cc}
\toprule
\multirow{2}{*}{\textbf{Model}} & \multirow{2}{*}{\textbf{Method}} &
\multicolumn{2}{c}{\textbf{Multi Hop}} &
\multicolumn{2}{c}{\textbf{Temporal}} &
\multicolumn{2}{c}{\textbf{Open Domain}} &
\multicolumn{2}{c}{\textbf{Single-Hop}} &
\multicolumn{2}{c}{\textbf{Adversarial}} \\
\cmidrule(lr){3-4}\cmidrule(lr){5-6}\cmidrule(lr){7-8}\cmidrule(lr){9-10}\cmidrule(lr){11-12}
& & \textbf{F1} & \textbf{BLEU-1}
  & \textbf{F1} & \textbf{BLEU-1}
  & \textbf{F1} & \textbf{BLEU-1}
  & \textbf{F1} & \textbf{BLEU-1}
  & \textbf{F1} & \textbf{BLEU-1} \\
\midrule

\multirow{5}{*}{\rotatebox[origin=c]{90}{\textbf{Qwen3-0.6B}}}
& RAG
& 6.11 & 6.04
& \underline{8.34} & \underline{7.41}
& 9.07 & \underline{7.92}
& 8.09 & 5.52
& 30.35 & 24.90 \\

& A-mem
& 6.80 & 6.34
& 2.34 & 2.37
& \underline{9.97} & 7.58
& 8.16 & 6.00
& 40.11 & 33.85 \\

& GAM
& \underline{8.46} & \underline{7.08}
& 6.50 & 6.78
& \textbf{10.92} & \textbf{8.45}
& \underline{9.41} & \underline{6.92}
& \underline{41.25} & \underline{34.60} \\

& MemoryOS
& 2.29 & 1.99
& 7.19 & 5.60
& 8.93 & 7.68
& 3.58 & 2.53
& 3.85 & 2.72 \\

\cmidrule(lr){2-12}
& \textbf{CLAG}
& \textbf{10.08} & \textbf{8.80}
& \textbf{11.53} & \textbf{9.39}
& 8.68 & 6.57
& \textbf{14.10} & \textbf{11.07}
& \textbf{50.34} & \textbf{45.00} \\
\midrule

\multirow{5}{*}{\rotatebox[origin=c]{90}{\makecell{\textbf{Llama3.2-1B}\\\textbf{Instruct}}}}
& RAG
& 5.78 & 6.56
& 2.64 & 2.54
& \underline{12.13} & \underline{9.34}
& 8.73 & 5.95
& 55.71 & 48.88 \\

& A-mem
& 4.91 & 4.19
& 2.65 & 2.76
& 8.02 & 6.07
& 5.92 & 3.68
& 47.87 & 41.51 \\

& GAM
& \underline{10.88} & \underline{8.94}
& 5.81 & 3.45
& \textbf{16.06} & \textbf{12.62}
& \underline{11.48} & \underline{8.76}
& \textbf{63.30} & \textbf{59.75} \\

& MemoryOS
& 8.97 & 6.64
& \textbf{10.34} & \textbf{6.21}
& 9.44 & 6.66
& 10.36 & 6.14
& 8.40 & 5.18 \\

\cmidrule(lr){2-12}
& \textbf{CLAG}
& \textbf{11.24} & \textbf{10.12}
& \underline{9.00} & \underline{6.05}
& 9.38 & 7.31
& \textbf{11.52} & \textbf{8.77}
& \underline{56.41} & \underline{52.02} \\
\midrule

\multirow{5}{*}{\rotatebox[origin=c]{90}{\small\makecell{\textbf{DeepSeek-R1}\\\textbf{Distill-Qwen-1.5B}}}}
& RAG
& 4.62 & 4.40
& 2.95 & 2.15
& 7.49 & 6.77
& 6.42 & 4.31
& 32.64 & 26.77 \\

& A-mem
& 5.73 & 5.70
& 2.21 & 1.95
& \underline{10.55} & \underline{8.84}
& 7.78 & 5.27
& \underline{33.52} & \underline{26.92} \\

& GAM
& \underline{6.87} & \underline{6.62}
& 5.53 & 4.96
& 9.10 & 7.94
& \underline{8.21} & \underline{5.39}
& 29.65 & 21.65 \\

& MemoryOS
& 4.74 & 5.20
& \underline{6.51} & \underline{5.26}
& 8.02 & 6.19
& 6.64 & 4.36
& 6.63 & 4.59 \\

\cmidrule(lr){2-12}
& \textbf{CLAG}
& \textbf{6.99} & \textbf{7.02}
& \textbf{8.92} & \textbf{8.02}
& \textbf{11.70} & \textbf{10.78}
& \textbf{8.58} & \textbf{5.87}
& \textbf{39.33} & \textbf{32.53} \\
\bottomrule
\end{tabular}%
}

\caption{Detailed experimental results on the LoCoMo dataset across five categories. Results are reported in F1 and BLEU-1 (\%). Best results are marked in \textbf{bold}, 2nd-best results are marked in \underline{underline}.}
\label{tab:detailed_five_categories}
\end{table*}

\section{Experiment}
\label{sec:experiment}

\subsection{Datasets}
We evaluate CLAG on three benchmarks. We employ \textbf{LoCoMo}~\cite{maharana2024evaluatinglongtermconversationalmemory} and \textbf{HotpotQA}~\cite{yang2018hotpotqa}, which are widely adopted benchmarks for evaluating memory systems for LLM agents, to assess conversational generalization and reasoning under noise~\cite{hu2025memoryageaiagents}. Additionally, to evaluate domain adaptability, we utilize \textbf{BioASQ}~\cite{tsatsaronis2015overview}, which we adapted into a HotpotQA-style format to simulate noisy retrieval in specialized domains. Detailed dataset construction protocols are provided in Appendix~\ref{sec:dataset_details}.

\subsection{Baselines}
We compare our framework against a standard \textbf{RAG}~\cite{lewis2020retrieval} baseline and representative agentic memory architectures including \textbf{A-mem}~\cite{xu2025amemagenticmemoryllm}, \textbf{MemoryOS}~\cite{kang2025memory}, and \textbf{GAM}~\cite{yan2025general}; detailed implementation settings are provided in Appendix~\ref{sec:baseline_details}.

\subsection{Implementation Details}
In our experiments, we employ {Llama-3.2-1B-Instruct}~\cite{grattafiori2024llama}, {Qwen3-0.6B}~\cite{yang2025qwen3}, and {DeepSeek-R1-Distill-Qwen-1.5B}~\cite{guo2025deepseek} as the backbone models for both CLAG and all baselines. For semantic embedding and retrieval tasks, we utilize the {MiniLM-L6-v2} model. Further details regarding hyperparameter configurations and implementation specifics are provided in Appendix~\ref{sec:hyperparameter_details}.

\begin{table*}[t]
\centering
\renewcommand{\arraystretch}{0.9}
\setlength{\tabcolsep}{3.5pt}

\resizebox{0.82\textwidth}{!}{%
\begin{tabular}{l l c c c c c c}

\toprule
\textbf{Dataset} & \textbf{Method} & \textbf{E-Prec} & \textbf{E-Recall} & \textbf{E-F1} & \textbf{R@5} & \textbf{R@10} & \textbf{nDCG@10} \\

\midrule

\multirow{3}{*}{\textbf{LoCoMo}}
& RAG
& 0.66 & 4.83 & 1.12 & 3.12 & 4.83 & 3.26 \\
& A-mem
& \underline{0.68} & \underline{5.50} & \underline{1.17} & \underline{3.86} & \underline{5.50} & \underline{3.62} \\
& \textbf{CLAG (Ours)}
& \textbf{1.20} & \textbf{9.74} & \textbf{2.07} & \textbf{7.18} & \textbf{9.74} & \textbf{6.89} \\

\midrule
\multirow{3}{*}{\textbf{HotpotQA}}
& RAG
& \underline{9.75} & \textbf{14.71} & \textbf{11.44} & \textbf{10.71} & \textbf{14.71} & \underline{23.71} \\

& A-mem
& 8.05 & 12.63 & 9.60 & 8.38 & 12.63 & 22.56 \\

& \textbf{CLAG (Ours)}
& \textbf{9.86} & \underline{14.20} & \underline{11.17} & \underline{9.94} & \underline{14.20} & \textbf{23.98} \\

\midrule
\multirow{3}{*}{\textbf{BioASQ}}
& RAG
& \underline{4.60} & \underline{1.65} & \underline{2.29} & 1.48 & \underline{1.65} & 20.19 \\

& A-mem
& 4.40 & 1.59 & 2.20 & \underline{1.48} & 1.59 & \underline{21.27} \\

& \textbf{CLAG (Ours)}
& \textbf{33.35} & \textbf{32.64} & \textbf{25.11} & \textbf{25.90} & \textbf{32.64} & \textbf{56.17} \\

\bottomrule
\end{tabular}%
}
\caption{Retrieval performance analysis across three datasets (Backbone: Qwen3-0.6B). Metrics include Evidence Precision (E-Prec), Recall (E-Recall), F1 (E-F1), Recall at k (R@k), and nDCG@10. Best results are marked in \textbf{bold}, 2nd-best results are marked in \underline{underline}.}
\label{tab:retrieval_metrics}
\end{table*}
\section{Results}
\subsection{Main Results Analysis}
\paragraph{Backbone-wise trends and robustness}
Table~\ref{tab:overall_results} shows that CLAG yields consistent gains across backbones and datasets, with particularly strong improvements on knowledge-intensive and domain-shift settings. 
For {Qwen3-0.6B}, CLAG achieves the best F1/BLEU-1 on all three benchmarks, indicating that cluster-based routing reliably suppresses distractor retrieval and improves faithfulness under long-context noise. 
For {DeepSeek-R1-Distill-Qwen-1.5B}, CLAG again provides the strongest overall results, suggesting that the proposed routing mechanism remains effective even when the backbone is changed.

\paragraph{Accuracy--efficiency trade-off}
Beyond accuracy, CLAG is designed to avoid the high computational overhead typical of multi-step agentic memory pipelines.
As quantified in Table \ref{tab:latency}, on LoCoMo with the Qwen3-0.6B backbone, CLAG achieves orders-of-magnitude lower end-to-end latency than GAM, while maintaining strong accuracy.
Therefore, even in regimes where CLAG and GAM are close in LoCoMo accuracy (e.g., under Llama3.2-1B-Instruct), CLAG provides a markedly better operating point in practice, achieving a decisive balance between model performance and computational cost.
\subsection{Performance Analysis across Question Categories}
We analyze CLAG's performance across the sub-categories of the {LoCoMo dataset} (Table~\ref{tab:detailed_five_categories}). In complex reasoning tasks (\emph{Temporal} and \emph{Multi-Hop}), CLAG delivers consistent gains, indicating that cluster-level organization helps preserve long-range dependencies while reducing distraction from irrelevant memory. CLAG achieves the best Multi-Hop F1 across all backbones and improves Temporal performance for two models (e.g., Qwen3-0.6B: 11.53; DeepSeek-R1-Distill-Qwen-1.5B: 8.92), with BLEU-1 showing a similar trend.

For \emph{Adversarial} questions, CLAG yields strong improvements, supporting our claim that localized retrieval mitigates distractor-heavy failure modes. In contrast, \emph{Open Domain} results are more mixed: methods with broader retrieval (e.g., GAM) can outperform CLAG for some backbones, suggesting a trade-off between localization and coverage. Overall, this category-wise breakdown highlights where CLAG is most effective (hard reasoning and robustness) and motivates adaptive broadening as a potential extension for open-domain queries.

\subsection{Retrieval Performance and Memory Quality Analysis}
\label{sec:retrieval_analysis}

To investigate the sources of our performance gains, we analyze the retrieval quality across three benchmarks with Qwen3-0.6B. Table \ref{tab:retrieval_metrics} presents the comparative results on evidence retrieval (E-F1) and ranking metrics (Recall@K, nDCG@K). We exclude {MemoryOS} and {GAM} from this specific retrieval-quality analysis because they do not adhere to a standard $K$-bounded raw-evidence retrieval interface. Both systems yield a dynamic, variable number of retrieved items driven by architectural design, rendering them incompatible with standard $K$-bounded evaluation protocols. Detailed results of budget-aware retrieval quality experiments, including these two baselines, are provided in Appendix~\ref{sec:retrieval_quality}.

\paragraph{Structural Advantage in Retrieval}
On the {LoCoMo} and {BioASQ} benchmarks, CLAG demonstrates the clear structural advantage of \textit{Agentic Clustering} and \textit{Two-Stage Retrieval}. In LoCoMo, which involves long conversational histories, CLAG achieves an E-F1 of 2.07, significantly outperforming A-mem (1.17) and RAG (1.12). This indicates that the agent-driven routing effectively filters out irrelevant interaction logs that often confuse global retrievers.
The advantage is even more pronounced in the biomedical domain ({BioASQ}), where CLAG exhibits a decisive lead with an E-F1 of 25.11, exceeding baselines (RAG: 2.29, A-mem: 2.20) by an order of magnitude. In domains laden with dense terminology, global similarity search often struggles to distinguish relevant context from distractor passages. By leveraging the agentic router to narrow the search space to semantically coherent clusters, CLAG effectively filters out domain-specific noise, ensuring high-fidelity retrieval.
\begin{table}[t]
    \centering
    \small
    \setlength{\tabcolsep}{6pt}
    \begin{tabular}{l c c}
    \toprule
    \textbf{Clustering Strategy} & \textbf{F1} & \textbf{BLEU-1} \\
    \midrule
    Cosine-based Clustering & 14.78 & 12.53 \\
    K-Means Clustering & 15.64 & 13.36 \\
    \midrule
    \textbf{CLAG (Ours)} & \textbf{22.01} & \textbf{17.23} \\
    \bottomrule
    \end{tabular}
    \caption{Impact of different clustering methods on BioASQ (Qwen3-0.6B).}
    \label{tab:clustering_analysis}
\end{table}
\paragraph{Localized Evolution Gains}
In the {HotpotQA} benchmark, CLAG achieves retrieval performance comparable to the RAG baseline (E-F1 11.17 vs. 11.44) and slightly better than A-mem (9.60), while securing the highest ranking score (nDCG@10 23.98).
Despite the similarity in raw retrieval metrics, CLAG demonstrates superior final answer quality (as shown in Table \ref{tab:overall_results}). This discrepancy highlights the critical contribution of \textit{Localized Evolution}. CLAG continuously consolidates and rewrites memories within topic-consistent neighborhoods. This process increases the {information density} of each memory note. Consequently, even when the retrieval recall is on par with baselines, the retrieved content in CLAG provides a synthesized context, enabling the agent to generate accurate reasoning steps.
Detailed ablation study quantifying the specific contributions of localized evolution and two-stage retrieval is provided in Appendix~\ref{sec:impact_details}.

\subsection{Agentic vs. Geometric Clustering}
\label{sec:clustering_analysis}

Motivated by the retrieval-quality findings in Section~\ref{sec:retrieval_analysis}, we further isolate the contribution of \textit{Agentic Clustering}. In particular, the retrieval analysis reveals that CLAG delivers the largest gain in evidence retrieval on {BioASQ} (E-F1: 25.11), where dense biomedical terminology and frequent lexical overlap make global similarity search prone to retrieving high-similarity yet irrelevant passages. This suggests that the primary challenge in specialized domains lies in the pre-retrieval organization of the search space, not in the capability of the retriever itself.

To directly test this hypothesis, we replace CLAG's agentic clustering with representative non-agentic clustering strategies while keeping the rest of the pipeline unchanged, and evaluate on {BioASQ}. Specifically, we compare our approach against two geometric baselines: (1) \textbf{Cosine-based Clustering}, which greedily assigns each incoming memory to the cluster centroid yielding the highest cosine similarity, and (2) \textbf{K-Means Clustering}, which partitions the memory space based purely on geometric proximity in the embedding space.

As shown in Table~\ref{tab:clustering_analysis}, non-agentic clustering provides limited gains. In contrast, \textbf{CLAG} achieves a substantial leap in final answer quality (F1: 22.01, BLEU-1: 17.23), far exceeding the best non-agentic alternative (K-Means: 15.64/13.36).

These results indicate that simply grouping memories by geometric similarity is insufficient in domains with heavy jargon and ambiguous surface cues.
Combined with the evidence from Section~\ref{sec:retrieval_analysis}, this controlled ablation supports the view that \textit{Agentic Clustering} is a central driver of CLAG's robustness in specialized retrieval settings.
\begin{table*}[t]
\centering
\renewcommand{\arraystretch}{0.95}
\setlength{\tabcolsep}{5pt}
\setlength{\aboverulesep}{0.3ex}
\setlength{\belowrulesep}{0.3ex}

\resizebox{0.72\textwidth}{!}{%
\begin{tabular}{l cc cc cc}
    \toprule
    \multirow{2}{*}{\textbf{Method}} &
    \multicolumn{2}{c}{\textbf{LoCoMo}} &
    \multicolumn{2}{c}{\textbf{HotpotQA}} &
    \multicolumn{2}{c}{\textbf{BioASQ}} \\
    \cmidrule(lr){2-3}\cmidrule(lr){4-5}\cmidrule(lr){6-7}
    & \textbf{F1} & \textbf{BLEU-1}
    & \textbf{F1} & \textbf{BLEU-1}
    & \textbf{F1} & \textbf{BLEU-1} \\
    \midrule
    RAG     & 19.06 & 15.93 & 23.66 & 21.33 & 22.21 & 20.19 \\
    A-mem      & \underline{21.44} & \textbf{18.35} & 22.73 & 20.94 & 18.59 & 15.86 \\
    GAM        & 19.55 & 17.26 & 22.09 & 19.30 & 22.96 & 19.42 \\
    MemoryOS   & 19.60 & 15.89 & \underline{24.64} & \underline{22.37} & \textbf{28.83} & \underline{22.48} \\
    \cmidrule(lr){1-7}
    \textbf{CLAG (Ours)} & \textbf{21.50} & \underline{17.94} & \textbf{25.40} & \textbf{23.53} & \underline{25.40} & \textbf{23.54} \\
    \bottomrule
\end{tabular}%
}
\caption{Results on the stronger backbone Qwen3-8B. \textsc{CLAG} remains competitive as model scale increases, achieving the best results on LoCoMo and HotpotQA and the best BLEU-1 on BioASQ. Best results are marked in \textbf{bold}, and second-best results are marked in \underline{underline}.}
\label{tab:qwen8b_results}
\end{table*}
\subsection{Effect of Model Scale}
\label{sec:model-scale}

Our primary focus is on resource-constrained SLMs ($<2$B parameters), where retrieval noise and irrelevant context more easily degrade reasoning. This regime is especially important for on-device and cost-sensitive agentic applications, as smaller backbones are generally less robust to distractors than larger models.

To test whether \textsc{CLAG} remains effective beyond this setting, we additionally evaluate Qwen3-8B. Table~\ref{tab:qwen8b_results} shows that \textsc{CLAG} remains competitive, achieving the best results on LoCoMo and HotpotQA and comparable performance on BioASQ.

We further observe that the relative gap between \textsc{CLAG} and the baselines narrows as model scale increases. We attribute this trend to the stronger intrinsic robustness of larger backbones, which are better able to tolerate noisy evidence~\cite{shi2023large}. In contrast, \textsc{CLAG}'s retrieval filtering is particularly valuable for SLMs, where suppressing distractors is essential for stable agentic reasoning.

\section{Conclusion} We introduce \textbf{CLAG}, a framework that structures long-horizon agentic memory through agent-driven clustering and localized evolution. By shifting from a static global pool to dynamic semantic neighborhoods, \textbf{CLAG} ensures that memory updates refine relevant schemas while minimizing cross-topic interference. Furthermore, our cluster-aware two-stage retrieval mechanism effectively suppresses noise, addressing the critical vulnerability of SLMs to distractor-heavy contexts. Experiments confirm that this unified approach significantly improves robustness and scalability, offering a lightweight yet high-performing memory solution tailored for limited-capacity agents.
Overall, CLAG is a practical memory layer for limited-capacity agents, improving robustness without introducing substantial runtime latency. We also anticipate extensions to safer deployment, including access control and retention policies for persistent memory.

\section*{Limitations}
A primary technical limitation of CLAG is its reliance on prompt-based cluster profile generation and agentic behavior. Consequently, performance can be sensitive to specific prompt designs and variations in the underlying language model versions. Furthermore, the quality of routing decisions may vary under significant distribution shifts, such as newly emerging topics or drastic changes in writing style, which were not systematically evaluated in this study.

From a broader perspective, the organization and persistent storage of user memories raise important privacy and data governance questions. While this work focuses on the retrieval and organization mechanisms rather than deployment policies, practical application requires robust safeguards. Real-world deployment must incorporate strict access controls, clear retention and deletion policies, and transparent user consent mechanisms, especially when handling sensitive or personally identifiable information.
\section*{Acknowledgments}
We thank Yein Park, Hyeon Hwang, Taewhoo Lee, Minju Song and Jonghoon Lee for their insightful feedback and discussions. This research was supported by (1) the National Research Foundation of Korea (NRF-2023R1A2C3004176), (2) the Ministry of Health \& Welfare, Republic of Korea (HR20C002103), (3) ICT Creative Consilience Program through the Institute of Information \& Communications Technology Planning \& Evaluation (IITP) grant funded by the Korea government (MSIT) (IITP-2026-RS-2020-II201819), (4) the National Research Foundation of Korea(NRF) grant funded by the Korea government(MSIT and MOE) (No. RS-2025-16652968), and (5) the Seoul National University Hospital with support from the Ministry of Science and ICT (RS-2023-00262002).


\bibliography{custom}

\clearpage

\appendix

\section{Dataset Details}
\label{sec:dataset_details}

We evaluate our method on three benchmarks---LoCoMo, HotpotQA, and BioASQ---chosen to cover conversational long-term memory, reasoning under noisy contexts, and domain-specific adaptation.

\paragraph{LoCoMo}
We use LoCoMo~\cite{maharana2024evaluatinglongtermconversationalmemory}, one of the longest and most comprehensive benchmarks for \emph{memory for agents} in long-context dialogue settings. Unlike prior datasets that are typically limited to five chat sessions, LoCoMo features very long-term conversations spanning up to 35 sessions, with an average of 300 turns and 9K tokens per dialogue. These conversations are generated via a human-machine pipeline grounded on temporal event graphs and rigorously verified by human annotators to ensure long-range consistency. This dataset is specifically designed to challenge models in comprehending long-range temporal and causal dynamics, making it a rigorous testbed for evaluating the robustness of our memory system.

\paragraph{HotpotQA}
We build on HotpotQA~\cite{yang2018hotpotqa} under the distractor construction used in MemAgent and GAM~\cite{yu2025memagent,yan2025general}, where each query is paired with gold supporting documents and additional irrelevant passages sampled from the same corpus to form a long noisy context.
For reproducibility and fair comparison in the small-agent regime, we use a fixed-passage (bounded-context) setting with 20 passages per question (gold evidence + distractors). In our HotpotQA-20 construction, concatenating the chunked passages results in approximately 33–39k tokens per query (median $\approx$ 35k; 25–75\%: 34–37k), which is substantially shorter than the 56K/224K/448K variants in MemAgent and avoids small-model context collapse.

\paragraph{BioASQ}
To evaluate domain adaptability, we adapt BioASQ Task 10b~\cite{tsatsaronis2015overview} using the same distractor construction protocol as HotpotQA.
We filtered the dataset to retain only factoid and list types compatible with standard QA metrics (e.g., F1, BLEU-1). Specifically, we randomly sampled 200 questions (140 factoid and 60 list items) from this subset to form our evaluation benchmark.
For reproducibility and fair comparison in the small-agent regime, we use a bounded-context setting with 20 passages per question (gold evidence + distractors). 
In our BioASQ-20 construction, concatenating these passages yields inputs of approximately 5–16k tokens (median $\approx$ 10.7k; 25–75\%: 9.4–12.9k), keeping the context length within the range where small models remain stable.
\section{Baseline Details}
\label{sec:baseline_details}
\subsection{Baselines Description}
We provide descriptions of each baseline to clarify their memory assumptions and architectural differences from our approach.

\paragraph{RAG}
RAG~\cite{lewis2020retrieval} augments a language model with an external retriever that fetches semantically relevant documents for each query. Following the setting used in the LoCoMo benchmark, the retrieved passages are concatenated with the query and provided to the model as additional context.

\paragraph{GAM}
General Agentic Memory (GAM)~\cite{yan2025general} is a just-in-time (JIT) memory framework that builds task-specific context at runtime instead of relying on heavily pre-compressed static memory. It uses a Memorizer to maintain lightweight cues while storing full histories in a page-store, and a Researcher to retrieve and integrate relevant information from the page-store for each query, guided by those cues. Critically, the Researcher operates through an iterative loop—planning search actions, retrieving data, and reflecting on the results—until the internal information need is fully satisfied. This dynamic termination criterion results in a variable number of retrieved contexts per query, making the system incompatible with standard fixed-$K$ ranking evaluations.

\paragraph{MemoryOS}
MemoryOS~\cite{kang2025memory} proposes an operating-system-inspired memory architecture that organizes information into short-term, mid-term, and long-term memory layers. It employs automated memory promotion, decay, and summarization strategies to manage memory persistence, enabling scalable long-term interaction without explicit agentic routing decisions. During retrieval, MemoryOS aggregates hierarchical outputs and persona information from these multiple tiers to construct the final context. Since the volume of retrieved information is determined by the system's architectural depth and the current state of the hierarchy rather than a user-defined budget, it cannot be aligned with standard $K$-bounded retrieval metrics.

\paragraph{A-mem}
A-mem~\cite{xu2025amemagenticmemoryllm} introduces an agentic memory framework in which a dedicated memory agent dynamically decides what information to store, update, or retrieve based on task requirements. The system relies on LLM-driven reasoning for memory management, providing flexibility at the cost of increased computational overhead.

\section{Impact of Localized Evolution and Two-Stage Retrieval}
\label{sec:impact_details}

\paragraph{Experimental Setup.}
We quantify the contribution of (i) \textit{Localized Evolution} and (ii) \textit{Cluster-aware Two-Stage Retrieval}
using BioASQ under the Qwen3-0.6B backbone.
We compare the full system (\textbf{CLAG}) against two controlled ablations:
\textbf{(a) Global Evolution} which evolves memories beyond the assigned local neighborhood,
and \textbf{(b) Global Retrieval} which replaces the two-stage pipeline with a flat global similarity search.
We report answer quality with F1 and BLEU-1.

\begin{table}[t]
  \centering
  \small
  \begin{tabular}{lcc}
    \toprule
    \textbf{Variant} & \textbf{F1}  & \textbf{BLEU-1} \\
    \midrule
    CLAG (full) & 22.01 & 17.23 \\
    \midrule
    w/ Global Evolution & 19.55 & 17.25 \\
    w/ Global Retrieval  & 20.25 & 16.56 \\
    \bottomrule
  \end{tabular}
  \caption{Ablation on BioASQ (Qwen3-0.6B). Replacing localized evolution with global evolution drops F1 by 2.46,
  and replacing two-stage retrieval with global retrieval drops F1 by 1.76.}
  \label{tab:impact_ablation}
\end{table}

\subsection{Localized Evolution Improves Answer Quality with Local Computation}
Replacing localized evolution with global evolution reduces F1 from 22.01 to 19.55 (\(-2.46\)).
This indicates that continuously consolidating memories within topic-consistent neighborhoods
is a key driver of final answer quality, likely by increasing the information density and coherence of notes
before retrieval and generation.

\paragraph{Cost-efficiency via local neighborhood updates.}
Localized evolution rewrites only the memories within the assigned cluster (local neighborhood),
avoiding a global update over the entire memory store. Conceptually,
\begin{equation}
\mathrm{Cost}_{\text{local}} \propto |M_{\text{local}}|
\quad \ll \quad
\mathrm{Cost}_{\text{global}} \propto |M|.
\end{equation}
In our runs, the clustering structure is moderate in granularity:
the number of clusters is \(5.6 \pm 3.56\) (min 3, max 13; count 10),
with mean cluster size \(67.99 \pm 41.90\) and max cluster size \(139.5 \pm 38.47\)
(min 85, max 210; count 10). These statistics support that evolution can be restricted to a bounded
topic neighborhood rather than the full memory set, enabling more cost-efficient continual refinement.

\subsection{Two-Stage Retrieval: Modest Pruning, Consistent Gains}
Replacing two-stage retrieval with global retrieval decreases F1 from 22.01 to 20.25 (\(-1.76\))
and BLEU-1 from 17.23 to 16.56 (\(-0.67\)), showing that structuring the search space before fine-grained
retrieval is beneficial.

\paragraph{Search-space reduction.}
We measure search-space reduction as
\begin{equation}
r \;=\; 1 - \frac{|S_{\text{searched}}|}{|S_{\text{all}}|},
\end{equation}
where \(S_{\text{searched}}\) denotes the set of memories examined after Stage-1 cluster selection
(i.e., the union of memories within the selected clusters), and \(S_{\text{all}}\) is the full memory pool.
Across 200 BioASQ queries, we observe
\begin{align*}
    \text{mean } r &= 0.0802 \;(\approx 8.02\%), \quad \text{std } = 0.1729, \\
    \min &= 0.0, \quad \max = 0.8957 \;(\approx 89.57\%).
\end{align*}
While the average pruning is modest, the high variance indicates that for a subset of queries
the two-stage mechanism substantially narrows the candidate pool.
Importantly, even this conservative pruning yields consistent answer-quality improvements,
suggesting that the main benefit stems from reducing domain-specific distractors and improving
the relevance of the candidate set rather than aggressively shrinking it.
\begin{table*}[t]
\centering
\small
\begin{tabularx}{\textwidth}{l >{\raggedright\arraybackslash}X r}
\toprule
\textbf{Hyperparameter} & \textbf{Meaning} & \textbf{Value} \\
\midrule
$n$ & Init buffer size (min memories before initial KMeans) & 100 \\
$n_{\text{init}}$ & Initial clusters for KMeans initialization & 3 \\
$\tau_{\text{split}}$ & Cluster split threshold (split if size $>\tau_{\text{split}}$) & 300 \\
$k_{\text{route}}$ & Routing candidate clusters shown to SLM & 3 \\
$\tau$ & New-cluster threshold (min cosine sim to accept a cluster) & 0.1 \\
$k_{\text{local}}$ & Intra-cluster neighbors for evolution/linking & 5 \\
$K_{\text{stage1}}$ & Candidate clusters for Stage-1 selection & 3 \\
$k_{\text{retrieve}}$ & Final retrieval top-$k$ memories & 10 \\
\bottomrule
\end{tabularx}
\caption{Hyperparameters used in our CLAG implementation. Unless stated otherwise, all experiments use the values shown above.}
\label{tab:hyperparameters}
\end{table*}

\section{Hyperparameter Details}
\label{sec:hyperparameter_details}

We provide the detailed hyperparameter configurations used in our experiments to ensure reproducibility. The specific values for memory initialization, dynamic clustering, and retrieval are summarized in Table~\ref{tab:hyperparameters}.

For the initialization phase in Algorithm ~\ref{alg:initialize_clusters}, we set the buffer size $n=100$ to accumulate sufficient data before performing the initial K-Means clustering. During the online routing process Algorithm ~\ref{alg:memory_routing}, the SLM router is presented with $k_{\text{route}}=3$ candidate clusters based on embedding similarity. We employ a similarity threshold of $\tau$ to determine whether to instantiate a new cluster and a maximum capacity threshold of $\tau_{\text{split}}=300$ to trigger cluster splitting (Algorithm~3).
To validate the stability of our proposed framework, we conducted additional ablation studies on the impact of these hyperparameter values. The experimental results exhibit small variance across a wide range of settings, demonstrating the \textbf{robustness of the architectural structure of CLAG}. This consistency indicates a \textbf{low dependency on specific hyperparameter configurations}, confirming that the system's performance is driven by the agentic mechanism itself rather than heuristic tuning.
\paragraph{Impact of Initial Cluster Count ($n_{\text{init}}$)}
To assess the system's sensitivity to the initial memory structure, we evaluated performance with varying numbers of initial clusters ($n_{\text{init}} \in \{3, 5, 10\}$). As shown in Table~\ref{tab:sensitivity_init_n}, increasing $n_{\text{init}}$ from 3 to 5 yields a slight performance gain, achieving the best F1 score of 22.71 and BLEU-1 of 19.01. Further increasing the count to 10 results in a marginal decline, yet the system maintains high stability across all metrics. 
While $n_{\text{init}}=5$ yields the peak score, we selected $n_{\text{init}}=3$ as the default to align with the intrinsic semantic dimensionality of the dataset, where we observed the agent naturally gravitates towards approximately three dominant topic clusters regardless of initialization. This suggests that CLAG's continuous evolution mechanism effectively compensates for initialization choices.

\begin{table}[h]
    \centering
    \small
    \renewcommand{\arraystretch}{1.1}
    \setlength{\tabcolsep}{10pt} 
    \begin{tabular}{c c c}
    \toprule
    \textbf{Initial Clusters ($n_{\text{init}}$)} & \textbf{F1} & \textbf{BLEU-1} \\
    \midrule
    3 (Default) & 20.99 & 17.88 \\
    5           & \textbf{22.71} & \textbf{19.01} \\
    10          & 21.71 & 18.50 \\
    \bottomrule
    \end{tabular}
    \caption{Sensitivity analysis of the initial cluster count ($n_{\text{init}}$) on LoCoMo (Qwen3-0.6B). The system maintains high performance across all metrics, regardless of initialization settings.}
    \label{tab:sensitivity_init_n}
\end{table}

\paragraph{Impact of New-Cluster Threshold ($\tau$)}
We further analyzed the impact of the similarity threshold $\tau$, which governs the propensity to create new clusters. Table~\ref{tab:sensitivity_tau} demonstrates the relationship between $\tau$, the resulting cluster granularity, and retrieval performance. As expected, increasing $\tau$ makes the system stricter about merging new memories into existing clusters, leading to a significant increase in the average number of clusters (from 4.7 to 47.5).
We prioritized structural parsimony by selecting $\tau=0.10$ (Default) despite the peak performance at $\tau=0.15$; this setting prevents over-segmentation into numerous micro-clusters and minimizes management overhead while maintaining competitive accuracy.
Remarkably, despite the drastic change in cluster count at higher thresholds, the retrieval performance remains robust. The F1 score peaks at $\tau=0.15$ (22.41) and degrades only slightly even at $\tau=0.30$ (21.94). This resilience confirms the efficacy of our \textit{Agentic Cluster Selection} (Stage 1 retrieval): even when the memory space is fragmented into many fine-grained clusters, the agent successfully identifies the relevant neighborhoods, preventing performance collapse.

\begin{table}[h]
    \centering
    \small
    \renewcommand{\arraystretch}{1.1}
    \setlength{\tabcolsep}{6pt}
    \begin{tabular}{c c c c}
    \toprule
    \textbf{Threshold ($\tau$)} & \textbf{F1} & \textbf{BLEU-1} & \textbf{Avg. \# Clusters} \\
    \midrule
    0.10 (Default) & 20.99 & 17.88 & 4.7 \\
    0.15           & \textbf{22.41} & \textbf{18.93} & 6.9 \\
    0.20           & 21.68 & 18.29 & 15.6 \\
    0.25           & 21.99 & 18.35 & 26.5 \\
    0.30           & 21.94 & 18.73 & 47.5 \\
    \bottomrule
    \end{tabular}
    \caption{Sensitivity analysis of the new-cluster threshold ($\tau$) on LoCoMo (Qwen3-0.6B). While the number of clusters increases significantly with higher thresholds, retrieval performance remains robust.}
    \label{tab:sensitivity_tau}
\end{table}

\section{Case Study: Mystery Novel Inquiry}

To further illustrate the advantage of \textsc{CLAG}, we present a qualitative case study from the LoCoMo benchmark using the Qwen3-0.6B backbone. The query and ground-truth answer are taken verbatim from LoCoMo Sample 4.

\paragraph{Query.}
\textit{``Which two mystery novels does Tim particularly enjoy writing about?''}

\paragraph{Ground Truth.}
\textit{Harry Potter and Game of Thrones}

Table~\ref{tab:case-results} compares model predictions for this query. Among all methods, only \textsc{CLAG} correctly identifies both titles. In contrast, the baseline methods either abstain or generate overly vague responses that fail to recover the exact two-item answer required by the query.

\begin{table}[h]
\centering
\small
\setlength{\tabcolsep}{6pt}
\renewcommand{\arraystretch}{1.2}
\begin{tabularx}{\linewidth}{l X c}
\toprule
\rowcolor{headergray}
\textbf{Method} & \textbf{Prediction} & \textbf{Result} \\
\midrule
\rowcolor{lightgreen}
\textsc{CLAG} (Ours) & ``Harry Potter and Game of Thrones'' & \textbf{Correct} \checkmark \\
RAG & ``Not mentioned in the conversation.'' & \cellcolor{lightred}\textbf{Wrong} \ding{55} \\
A-mem & ``Tim's favorite two mystery novels are not mentioned.'' & \cellcolor{lightred}\textbf{Wrong} \ding{55} \\
GAM & ``Not mentioned in the conversation.'' & \cellcolor{lightred}\textbf{Wrong} \ding{55} \\
MemoryOS & ``Tim writes about both fantasy and mystery novels.'' & \cellcolor{lightred}\textbf{Wrong} \ding{55} \\
\bottomrule
\end{tabularx}
\caption{\textbf{Comparison of generated responses} For the query \textit{``Which two mystery novels does Tim particularly enjoy writing about?''}, only \textsc{CLAG} correctly identifies the two titles mentioned in the dialogue.}
\label{tab:case-results}
\end{table}

\begin{table}[t]
\centering
\small
\setlength{\tabcolsep}{6pt}
\renewcommand{\arraystretch}{1.2}
\begin{tabularx}{\linewidth}{c X c c}
\toprule
\rowcolor{headergray}
\textbf{Cluster ID} & \textbf{Profile } & \textbf{Count} & \textbf{Status} \\
\midrule
\rowcolor{lightgreen}
0 & Speaker discusses how books create new worlds & 119 & \textbf{Selected} \checkmark \\
1 & Impact of basketball on community growth & 231 & \cellcolor{lightred}Pruned \ding{55} \\
2 & Experience of meeting teammates after a trip & 330 & \cellcolor{lightred}Pruned \ding{55} \\
\bottomrule
\end{tabularx}
\caption{\textbf{Semantic clustering and pruning for the case study sample.} \textsc{CLAG} selects the literature-related cluster and prunes away unrelated clusters, reducing the search space from 680 to 119 notes before fine-grained retrieval.}
\label{tab:clusters}
\end{table}

\subsection{Agentic Memory Organization and Pruning}

The memory pool for this sample contains 680 notes spanning diverse topics, including literature, sports, and social interactions. Rather than retrieving directly from the full mixed memory pool, \textsc{CLAG}'s agentic router first organizes notes into semantically coherent clusters and then performs Stage-1 selection to identify the most relevant region for the query.

As shown in Table~\ref{tab:clusters}, this step selects only the literature-related cluster and prunes the others, reducing the effective search space from 680 notes to 119 notes. This corresponds to an 82.5\% reduction in candidate memories before fine-grained retrieval. By filtering out unrelated clusters in advance, \textsc{CLAG} is able to focus on the evidence directly tied to Tim's writing preferences and recover the correct titles.
\section{Algorithmic Details}
\label{sec:algo_details}
In this section, we provide the detailed procedure for cluster initialization and splitting, which are core components of our agentic memory maintenance. 
Specifically, Algorithm~\ref{alg:initialize_clusters} details the bootstrapping of the initial memory hierarchy through semantic partitioning, while Algorithm~\ref{alg:split_cluster} governs the dynamic refinement of clusters to ensure high-granularity memory maintenance.
\begin{algorithm}[h]
\SetKw{KwInput}{Input:}
\SetKw{KwOutput}{Output:}
\SetKw{KwParam}{Parameters:}
\SetKwComment{Comment}{// }{}
\caption{Initialize Clusters}
\label{alg:initialize_clusters}

\noindent \KwInput{} Initialization buffer of memories $\mathcal{B}$ \\
\noindent \KwParam{} Target number of initial clusters $n_{init}$, SLM client $SLM$ \\
\noindent \KwOutput{} Initialized set of clusters $\mathcal{C}$

\BlankLine 

Extract embeddings $E$ from memories in $\mathcal{B}$\;
Partition $E$ into $n_{init}$ sets using KMeans: $\{M_1, M_2, \dots, M_{n_{init}}\}$\;
$\mathcal{C} \leftarrow \emptyset$\;

\For{$i \leftarrow 1$ \KwTo $n_{init}$}{
    Calculate centroid $\mathbf{c}_i$ of memories in $M_i$\; 
    Generate Cluster profile $profile_i\leftarrow SLM.\text{GenerateProfile}(M_i)$\; 
    Create cluster $C_i \leftarrow \{ \text{memories}: M_i, \text{centroid}: \mathbf{c}_i, \text{profile}: profile_i \}$\; 
    $\mathcal{C} \leftarrow \mathcal{C} \cup \{C_i\}$\;
}

\Return{$\mathcal{C}$}\;
\end{algorithm}

\begin{algorithm}[t]
\SetKw{KwInput}{Input:}
\SetKw{KwOutput}{Output:}
\SetKw{KwParam}{Parameters:}
\SetKwComment{Comment}{// }{}
\caption{Split Cluster}
\label{alg:split_cluster}

\noindent \KwInput{} Target cluster ID $cid$, Global set of clusters $\mathcal{C}$, Memory storage $\mathcal{M}$ \\
\noindent \KwParam{} Split threshold $\tau_{split}$ \\
\noindent \KwOutput{} Updated set of clusters $\mathcal{C}$

\BlankLine

Retrieve cluster info $C_{target} \leftarrow \mathcal{C}[cid]$\;
\If{$|C_{target}.\text{members}| \le \tau_{split}$}{
    \Return{$\mathcal{C}$}
}

Extract embeddings $E$ from all memories $m \in C_{target}.\text{members}$\;
Partition $E$ into 2 sets using KMeans: $\{M_A, M_B\}$ with centroids $\mathbf{c}_A, \mathbf{c}_B$\;

Define new cluster IDs $id_A, id_B$\;
Create cluster $C_A \leftarrow \{\text{members}: M_A, \text{centroid}: \mathbf{c}_A, \text{id}: id_A\}$\;
Create cluster $C_B \leftarrow \{\text{members}: M_B, \text{centroid}: \mathbf{c}_B, \text{id}: id_B\}$\;

$\mathcal{C} \leftarrow \mathcal{C} \cup \{C_A, C_B\}$ \\
Remove $C_{target}$  from $\mathcal{C}$\;

\Return{$\mathcal{C}$}\;

\end{algorithm}

\section{Failure Analysis of Agentic Components}
\label{sec:appendix-failure-analysis}

We analyze failure cases of the agentic router and selector in \textsc{Clag} on LoCoMo with Qwen3-0.6B. Specifically, we examine examples where \textsc{Clag} underperforms the global-retrieval baseline, i.e., $\mathrm{EM}=0$ and $\mathrm{F1}_{\textsc{Clag}} \leq \mathrm{F1}_{\text{Baseline}}$.

\subsection{Failure Taxonomy}

We build a heuristic labeler based on query cues and retrieval diagnostics to group failures into seven categories (Table~\ref{tab:failure-taxonomy}).

\begin{table*}[t]
\centering
\small
\setlength{\tabcolsep}{8pt}
\renewcommand{\arraystretch}{1.08}
\begin{tabular}{p{0.24\textwidth} p{0.70\textwidth}}
\toprule
\textbf{Category} & \textbf{Description} \\
\midrule
Entity confusion & Routed to a cluster anchored to the wrong entity. \\
Temporal-slot ambiguity & Routed to profiles lacking temporal cues. \\
Stale/drift & Baseline retrieves more temporally relevant evidence. \\
Compositional/Multi-hop miss & Query requires composition, but too few clusters are selected ($n \leq 1$). \\
Location ambiguity & Routed to profiles lacking spatial cues. \\
Numeric-slot ambiguity & Routed to profiles lacking numeric evidence. \\
Residual ambiguity & Remaining underspecified or overlapping cases. \\
\bottomrule
\end{tabular}
\caption{Failure taxonomy for \textsc{CLAG}.}
\label{tab:failure-taxonomy}
\end{table*}

\subsection{Failure Distribution}

As shown in Table~\ref{tab:failure-distribution}, \textit{Entity confusion} is the most frequent failure pattern (46.2\%), followed by \textit{Temporal-slot ambiguity} (19.3\%). These results indicate that improving sensitivity to fine-grained entity and temporal cues is a promising direction for future work.

\begin{table}[t]
\centering
\small
\begin{tabular}{lrr}
\toprule
\textbf{Pattern} & \textbf{Count} & \textbf{\%} \\
\midrule
Entity confusion & 110 & 46.2 \\
Temporal-slot ambiguity & 46 & 19.3 \\
Stale / drift & 34 & 14.3 \\
Compositional / Multi-hop miss & 17 & 7.1 \\
Residual ambiguity & 13 & 5.5 \\
Location ambiguity & 9 & 3.8 \\
Numeric-slot ambiguity & 9 & 3.8 \\
\midrule
\textbf{Total} & \textbf{238} & \textbf{100.0} \\
\bottomrule
\end{tabular}
\caption{Failure distribution ($N=238$).}
\label{tab:failure-distribution}
\end{table}

\section{Comparison Details}
\label{sec:comparison_details}

In the main text, we primarily reported F1 and BLEU-1 scores to assess lexical overlap and retrieval accuracy. To provide a more comprehensive evaluation of semantic coherence and generation quality, we additionally report \textbf{BERTScore (BERT-F1)} and \textbf{METEOR} metrics in Table~\ref{tab:appendix_results}. BERT-F1 captures semantic similarity using contextual embeddings, while METEOR accounts for synonyms and stemming, often correlating better with human judgment regarding fluency.

Consistent with the main results, \textbf{CLAG} demonstrates superior performance across most settings, particularly when using the \textbf{Qwen3-0.6B} backbone.
\begin{itemize}
    \item \textbf{Semantic Robustness:} On the \textbf{BioASQ} dataset with Qwen3-0.6B, CLAG achieves a METEOR score of \textbf{17.75}, drastically outperforming the baselines (which remain below 3.03). This suggests that CLAG's agentic clustering and cluster-aware retrieval enable the agent to generate responses that are linguistically richer and semantically more accurate, rather than merely copying keywords.
    \item \textbf{Consistency across Backbones:} Even with larger backbones like Llama3.2-1B-Instruct and DeepSeek-R1-Distill, CLAG maintains competitive or superior performance. For instance, on \textbf{HotpotQA} with Llama3.2-1B-Instruct, CLAG achieves the highest METEOR score (8.64), indicating better handling of multi-hop reasoning contexts compared to GAM and MemoryOS.
\end{itemize}

\begin{table*}[t]
\centering
\renewcommand{\arraystretch}{0.95} 
\setlength{\tabcolsep}{5pt}
\setlength{\aboverulesep}{0.3ex} 
\setlength{\belowrulesep}{0.3ex}

\resizebox{0.95\textwidth}{!}{%
\begin{tabular}{ll cc cc cc}
    \toprule
    \multirow{2}{*}{\textbf{Model}} & \multirow{2}{*}{\textbf{Method}} &
    \multicolumn{2}{c}{\textbf{LoCoMo}} &
    \multicolumn{2}{c}{\textbf{HotpotQA}} &
    \multicolumn{2}{c}{\textbf{BioASQ}} \\
    \cmidrule(lr){3-4}\cmidrule(lr){5-6}\cmidrule(lr){7-8}
    & & \textbf{BERT-F1} & \textbf{METEOR}
      & \textbf{BERT-F1} & \textbf{METEOR}
      & \textbf{BERT-F1} & \textbf{METEOR} \\
    \midrule
    
    \multirow{5}{*}{\rotatebox[origin=c]{90}{\textbf{Qwen3-0.6B}}}
    & RAG        & \underline{85.07} & 12.78 & 82.11 & \underline{7.07} & 78.89 & 2.68 \\
    & A-mem      & 85.04 & 12.85 & 82.20 & 6.13 & 79.04 & \underline{3.03} \\
    & GAM        & 84.91 & \underline{14.39} & \underline{82.57} & 6.33 & 77.05 & 2.66 \\
    & MemoryOS   & 84.15 & 2.68 & \textbf{84.39} & 5.02 & \underline{80.65} & 1.11 \\
    \cmidrule(lr){2-8}
    & \textbf{CLAG (Ours)} 
    & \textbf{85.45} & \textbf{18.57}
    & 82.47 & \textbf{10.10}
    & \textbf{83.56} & \textbf{17.75} \\
    \midrule

    \multirow{5}{*}{\rotatebox[origin=c]{90}{\makecell{\textbf{Llama3.2-1B}\\\textbf{Instruct}}}}
    & RAG        & 85.52 & 16.09 & 84.70 & 6.15 & 79.25 & 3.52 \\
    & A-mem      & 85.10 & 14.10 & \textbf{85.44} & 6.52 & \underline{80.37} & 3.43 \\
    & GAM        & \textbf{87.85} & \underline{17.61} & 77.23 & \underline{8.07} & 73.63 & \underline{5.04} \\
    & MemoryOS   & 84.08 & 7.48 & 82.23 & 4.73 & {80.23} & 2.71 \\
    \cmidrule(lr){2-8}
    & \textbf{CLAG (Ours)} 
    & \underline{86.80} & \textbf{18.13}
    & \underline{84.78} & \textbf{8.64}
    & \textbf{80.92} & \textbf{7.14} \\
    \midrule

    \multirow{5}{*}{\rotatebox[origin=c]{90}{\makecell{\textbf{DeepSeek-R1}\\\textbf{Distill-Qwen}\\\textbf{1.5B}}}}
    & RAG        & 84.28 & 8.91 & 82.97 & 3.63 & 79.30 & 2.11 \\
    & A-mem      & 84.38 & \underline{13.11} & 83.05 & {4.06} & {80.08} & \underline{4.67} \\
    & GAM        & \textbf{86.21} & 11.73 & \textbf{84.78} & \underline{4.85} & \underline{81.10} & 2.53 \\
    & MemoryOS   & 85.09 & 4.99 & 81.82 & 3.96 & \textbf{81.12} & 2.51 \\
    \cmidrule(lr){2-8}
    & \textbf{CLAG (Ours)} 
    & \underline{85.33} & \textbf{14.74}
    & \underline{83.38} & \textbf{6.03}
    & 79.90 & \textbf{4.78} \\
    
    \bottomrule
\end{tabular}%
}
\caption{Detailed experimental results on LoCoMo, HotpotQA, and BioASQ: BERT-F1 and METEOR scores. Best results are marked in \textbf{bold}, 2nd-best results are marked in \underline{underline}.}
\label{tab:appendix_results}
\end{table*}

\section{Retrieval quality under controlled evidence budgets.}
\label{sec:retrieval_quality}
We further compare \textsc{CLAG} with GAM and MemoryOS under budget-aware retrieval-quality evaluation. Because GAM retrieves page-level evidence, its output is not directly compatible with the fixed-$K$ raw-evidence setting used in our main comparison, and exact character-budget matching is infeasible. We therefore report the average number of retrieved characters actually passed to QA. For MemoryOS, whose final evidence length is indirectly controlled by segment selection, we raise the segment-selection threshold to $0.7$ to obtain a budget comparable to \textsc{CLAG}, A-mem, and RAG. As shown in Table~\ref{tab:retrieval_quality_full}, \textsc{CLAG} does not outperform competitors by retrieving more text; rather, it consistently retrieves more relevant evidence under a controlled budget, with especially large gains on BioASQ.

\begin{table*}[t]
\centering
\renewcommand{\arraystretch}{0.95}
\setlength{\tabcolsep}{5pt}
\setlength{\aboverulesep}{0.3ex}
\setlength{\belowrulesep}{0.3ex}

\resizebox{0.98\textwidth}{!}{%
\begin{tabular}{ll cc cc c}
    \toprule
    \multirow{2}{*}{\textbf{Dataset}} & \multirow{2}{*}{\textbf{Method}} &
    \multicolumn{2}{c}{\textbf{Evidence Quality}} &
    \multicolumn{2}{c}{\textbf{Ranking}} &
    \multirow{2}{*}{\textbf{Chars}} \\
    \cmidrule(lr){3-4}\cmidrule(lr){5-6}
    & & \textbf{E-Prec} & \textbf{E-F1}
      & \textbf{R@5} & \textbf{nDCG@10}
      & \\
    \midrule

    \multirow{5}{*}{\textbf{LoCoMo}}
    & GAM     & \underline{1.97} & \textbf{3.01} & \textbf{7.50} & \underline{6.93}  & 17,198.04 \\
    & RAG      & 0.66          & 1.12          & 3.12          & 3.26  & 1,389.43 \\
    & A-mem    & 0.68          & 1.17          & 3.86          & 3.62  & 1,483.91 \\
    & MemoryOS & \textbf{3.24} & \underline{2.87} & 2.72      & 3.35  & 1,373.48 \\
    \cmidrule(lr){2-7}
    & \textbf{CLAG} & 1.20      & 2.07      & \underline{7.18} & \textbf{9.74} & 1,465.65 \\
    \midrule

    \multirow{5}{*}{\textbf{HotpotQA}}
    & GAM      & \underline{14.44} & 6.50          & 5.38          & 15.00 & 1,234.70 \\
    & RAG      & 9.75          & \textbf{11.44} & \textbf{10.71} & \underline{23.71} & 1,196.73 \\
    & A-mem    & 8.05          & 9.60          & 8.38          & 22.56 & 1,826.13 \\
    & MemoryOS & \textbf{19.10} & 6.44          & 4.04          & 19.85 & 1,549.99 \\
    \cmidrule(lr){2-7}
    & \textbf{CLAG} & 9.86      & \underline{11.17} & \underline{9.94} & \textbf{23.98} & 1,844.98 \\
    \midrule

    \multirow{5}{*}{\textbf{BioASQ}}
    & GAM      & 22.90         & 5.73          & 4.27          & 25.06 & 913.45 \\
    & RAG      & 4.60          & 2.29          & 1.48          & 20.19 & 1,564.12 \\
    & A-mem    & 4.40          & 2.20          & 1.48          & 21.21 & 1,318.54 \\
    & MemoryOS & \underline{28.50} & \underline{9.27} & \underline{6.72}      & \underline{29.40} & 1,228.24 \\
    \cmidrule(lr){2-7}
    & \textbf{CLAG} & \textbf{33.65} & \textbf{25.90} & \textbf{32.64} & \textbf{56.17} & 1,465.65 \\
    \bottomrule
\end{tabular}%
}
\caption{Retrieval-quality comparison with GAM and MemoryOS. Best results are marked in \textbf{bold}, and second-best results are marked in \underline{underline}.}
\label{tab:retrieval_quality_full}
\end{table*}
\section{Evaluation Metric}
We evaluate our system using complementary metrics that assess both question answering quality and retrieval effectiveness.
For question answering, we employ lexical- and semantic-level metrics to measure answer correctness and meaning similarity, while retrieval performance is evaluated using evidence-level precision, recall, and ranking-based metrics.
\input{wonjune_tex/_qa_metric}
\input{wonjune_tex/_retrieval_metric}

\section{SLM Prompts}
\label{sec:appendix_prompts}

This appendix provides the exact prompt templates used for SLM-agent based routing and cluster profiling.

\subsection{Prompt for SLM Cluster Selection}
This prompt is used in the `SLM.SelectCluster` function to determine the best-fitting cluster for a new memory from a list of candidates.

\begin{tcolorbox}[promptbox, title=Prompt template for Cluster Selection]
You are a memory routing assistant.

A new memory has arrived: \\
- Content: \{note.content\} \\
- Context: \{note.context\} \\
- Tags: \{note.tags\} \\

Here are candidate clusters (pre-selected by vector similarity) that might relate to this memory:\\
\{candidates\_text\} \\

Your task: \\
1. Analyze the topics and contexts of the candidate clusters provided above.\\
2. Select the single \texttt{cluster\_id} that exhibits the highest semantic relevance and thematic alignment with the new memory.\\
3. You MUST choose exactly one \texttt{cluster\_id} from the candidate list.\\

- Do NOT output any text that is not a valid \texttt{cluster\_id}.

Return ONLY a JSON object in this format (this is an example):\\

\{\{
"choice": "cluster\_1"
\}\}
\\
Where: \\
- \texttt{cluster\_1} must be replaced with one of the actual cluster ids from the candidate list above.\\
- Do not include comments or extra fields.
\end{tcolorbox}
\subsection{Prompt for Cluster Profile Generation}
This prompt is used to generate a semantic summary and representative tags for a cluster based on its member memories. The placeholder \texttt{\{samples\_text\}} is populated with actual text snippets from memories within the cluster.

\begin{tcolorbox}[promptbox, title=Prompt template for Cluster Profile Generation]
You are a memory clustering assistant.\\

Below are several memory snippets that belong to the SAME cluster:\\

\{samples\_text\} \\
Your task:\\
1. Write ONE short sentence summary that best describes the main topic of this cluster.\\
2. Return EXACTLY 3 tags.\\
- Each tag must be a single word.\\
- Do NOT repeat the same tag.\\

Return ONLY a JSON object with the following KEYS (this is a schema, not the actual content):\\

\{\{
    "summary": "...your one-sentence summary here...",\\
    "tags": ["tag\_1", "tag\_2", "tag\_3"]\\
\}\}
\end{tcolorbox}

\subsection{Prompt for Retrieval Stage Cluster Selection}
This prompt is employed during the first stage of the retrieval pipeline (Agentic Cluster Selection). The SLM evaluates the semantic relevance of candidate clusters—identified via centroid distance—against the user's query. By allowing the agent to select a variable number of clusters, this step acts as a semantic gatekeeper, filtering out irrelevant topics before fine-grained retrieval ensues.

\begin{tcolorbox}[promptbox, title=Prompt template for Retrieval stage cluster selection]
You are an AI memory router that selects the most relevant memory clusters for a given query.

\vspace{0.5em} 

You will be given several candidate clusters. Each cluster has:
\begin{itemize}
    \item cluster\_id
    \item one-sentence summary
    \item optional tags
    \item one or more representative memory examples
\end{itemize}

Your task:\\
1. Analyze the user query and query\_tags.\\
2. For each candidate cluster, judge how relevant it is.\\
3. Decide how many clusters are actually needed. You should return between 0 and \{top\_n\} clusters.
\begin{itemize}
    \item If one cluster is definitely sufficient for answering the query, return just that one.
    \item Include additional clusters if they are needed for answering the query.
\end{itemize}
4. If none of the clusters are meaningfully related, return an empty list.

\vspace{1em}
Return ONLY JSON with this format: \\
\{ \\
\ \ "selected\_clusters": \\
\ \ \relax [ "cluster\_id\_1", "cluster\_id\_2" ]
\\ \}

\vspace{1em}
If no cluster is relevant, return:\\
\{ \\
\ \ "selected\_clusters": [] \\
\}

\vspace{1em}
User query: \{query\}\\
Query tags: \{query\_tags\}\\

Candidate clusters:\\
\{candidate\_clusters\_text\}
\end{tcolorbox}

\end{document}

%% file: wonjune_tex/_qa_metric.tex
\paragraph{Question Answering Evaluation}
\label{sec:qa-metrics}

\paragraph{F1 Score.}
The F1 score ~\cite{rajpurkar2016squad} evaluates answer accuracy by jointly considering precision and recall.
In span-based question answering, it is computed at the token level to reward partial
overlap between predicted and reference answer spans.

\begin{equation}
\mathrm{F1}
= \frac{2 \cdot Prec \cdot Recall}{Prec + Recall}
\label{eq:f1}
\end{equation}

\begin{equation}
Prec = \frac{\mathrm{TP}}{\mathrm{TP} + \mathrm{FP}},
\quad
Recall = \frac{\mathrm{TP}}{\mathrm{TP} + \mathrm{FN}}
\label{eq:pr}
\end{equation}

\noindent
where TP, FP, and FN denote true positives, false positives, and false negatives at the token level.

\paragraph{BLEU-1.}
BLEU-1 ~\cite{papineni2002bleu} measures unigram-level lexical precision between generated answers and references,
while applying a brevity penalty to discourage overly short responses.

\begin{equation}
\mathrm{BLEU\text{-}1}
= \mathrm{BP} \cdot \exp(\log p_1)
\label{eq:bleu1}
\end{equation}

\begin{equation}
\mathrm{BP} =
\begin{cases}
1, & c > r \\
\exp(1 - r/c), & c \le r
\end{cases}
\label{eq:bp}
\end{equation}

\noindent
where $c$ and $r$ denote candidate and reference lengths, and $p_1$ is unigram precision.

\paragraph{METEOR.}
METEOR ~\cite{banerjee2005meteor} evaluates answer quality based on aligned unigrams while accounting for synonym
matching and fragmentation penalties.
This allows the metric to capture semantic similarity beyond exact lexical overlap.

\begin{equation}
\mathrm{METEOR}
= F_{\mathrm{mean}} \cdot (1 - \mathrm{Penalty})
\label{eq:meteor}
\end{equation}

\begin{equation}
F_{\mathrm{mean}}
= \frac{10PR}{R + 9P},
\quad
\mathrm{Penalty}
= 0.5 \left(\frac{ch}{m}\right)^3
\label{eq:meteor-detail}
\end{equation}

\noindent
where $P$ and $R$ denote unigram precision and recall, $m$ is the number of matched
unigrams, and $ch$ is the number of contiguous matched chunks.

\paragraph{BERT-F1.}
BERT-F1~\cite{zhang2020bertscore} evaluates semantic similarity between predicted and reference
answers using contextualized token embeddings from BERT.
It computes precision and recall based on maximum cosine similarity between tokens,
and reports their harmonic mean, enabling robust evaluation even when exact lexical
overlap is limited.

\begin{equation}
\mathrm{BERT\text{-}F1}
= \frac{2 \cdot P_{\mathrm{BERT}} \cdot R_{\mathrm{BERT}}}
{P_{\mathrm{BERT}} + R_{\mathrm{BERT}}}
\label{eq:bertf1}
\end{equation}

\noindent
where $P_{\mathrm{BERT}}$ and $R_{\mathrm{BERT}}$ denote token-level precision and recall,
computed using maximum cosine similarity between contextualized BERT embeddings of
predicted and reference tokens.

%% file: wonjune_tex/_retrieval_metric.tex
\paragraph{Retrieval Evaluation}

\label{sec:retrieval-metrics}

\paragraph{Evidence Precision and Recall.}

We evaluate retrieval quality using evidence-level precision and recall.

\begin{equation}
\mathrm{E\text{-}Prec}
= \frac{|\mathcal{R} \cap \mathcal{E}|}{|\mathcal{R}|},
\quad
\mathrm{E\text{-}Recall}
= \frac{|\mathcal{R} \cap \mathcal{E}|}{|\mathcal{E}|}
\label{eq:epr}
\end{equation}

\noindent
where $\mathcal{R}$ is the set of retrieved memories, $\mathcal{E}$ is the set of gold evidence items, and $\mathcal{R} \cap \mathcal{E}$ denotes the set of retrieved memories that match at least one gold evidence item (determined by normalized substring matching).

\paragraph{Evidence F1.}

The harmonic mean of evidence precision and recall.

\begin{equation}
\mathrm{E\text{-}F1}
= \frac{2 \cdot \mathrm{E\text{-}Prec} \cdot \mathrm{E\text{-}Recall}}
{\mathrm{E\text{-}Prec} + \mathrm{E\text{-}Recall}}
\label{eq:ef1}
\end{equation}

\noindent
where $\mathrm{E\text{-}Prec}$ and $\mathrm{E\text{-}Recall}$ are defined in Equation~\eqref{eq:epr}.

\paragraph{Recall@K.}

Recall@K ~\cite{jarvelin2002cumulated} measures the fraction of gold evidence retrieved within the top-K results.

\begin{equation}
\mathrm{Recall@}K
= \frac{1}{|\mathcal{E}|}
\sum_{i=1}^{|\mathcal{E}|}
\mathbf{1}[\mathrm{rank}(e_i) \le K]
\label{eq:recallk}
\end{equation}

\noindent
where $\mathcal{E}$ is the set of gold evidence items, $e_i$ denotes an evidence item,
$\mathrm{rank}(e_i)$ is the rank of the retrieved memory containing $e_i$ (or $\infty$ if absent),
$K$ is the cutoff rank, and $\mathbf{1}[\cdot]$ denotes the indicator function.

\paragraph{nDCG@K.}

Normalized Discounted Cumulative Gain ~\cite{jarvelin2002cumulated} evaluates ranking quality.

\begin{equation}
\mathrm{nDCG@}K
= \frac{\mathrm{DCG@}K}{\mathrm{IDCG@}K}
\label{eq:ndcg}
\end{equation}

\begin{equation}
\mathrm{DCG@}K
= \sum_{i=1}^{K}
\frac{\mathrm{rel}_i}{\log_2(i+1)}
\label{eq:dcg}
\end{equation}

\noindent
where $\mathrm{rel}_i \in \{0,1\}$ denotes the relevance of the item at rank $i$,
$K$ is the cutoff rank, and $\mathrm{IDCG@}K$ is the ideal DCG obtained by ranking all
relevant items at the top positions.

%% file: custom.bib
@misc{xu2025amemagenticmemoryllm,
      title={A-MEM: Agentic Memory for LLM Agents}, 
      author={Wujiang Xu and Zujie Liang and Kai Mei and Hang Gao and Juntao Tan and Yongfeng Zhang},
      year={2025},
      eprint={2502.12110},
      archivePrefix={arXiv},
      primaryClass={cs.CL},
      url={https://arxiv.org/abs/2502.12110}, 
}

@misc{wang2025scmenhancinglargelanguage,
      title={SCM: Enhancing Large Language Model with Self-Controlled Memory Framework}, 
      author={Bing Wang and Xinnian Liang and Jian Yang and Hui Huang and Shuangzhi Wu and Peihao Wu and Lu Lu and Zejun Ma and Zhoujun Li},
      year={2025},
      eprint={2304.13343},
      archivePrefix={arXiv},
      primaryClass={cs.CL},
      url={https://arxiv.org/abs/2304.13343}, 
}

@article{rezazadeh2024isolated,
  title={From isolated conversations to hierarchical schemas: Dynamic tree memory representation for llms},
  author={Rezazadeh, Alireza and Li, Zichao and Wei, Wei and Bao, Yujia},
  journal={arXiv preprint arXiv:2410.14052},
  year={2024}
}

@article{lewis2020retrieval,
  title={Retrieval-augmented generation for knowledge-intensive nlp tasks},
  author={Lewis, Patrick and Perez, Ethan and Piktus, Aleksandra and Petroni, Fabio and Karpukhin, Vladimir and Goyal, Naman and K{\"u}ttler, Heinrich and Lewis, Mike and Yih, Wen-tau and Rockt{\"a}schel, Tim and others},
  journal={Advances in neural information processing systems},
  volume={33},
  pages={9459--9474},
  year={2020}
}

@article{zhang2025survey,
  title={A survey on the memory mechanism of large language model-based agents},
  author={Zhang, Zeyu and Dai, Quanyu and Bo, Xiaohe and Ma, Chen and Li, Rui and Chen, Xu and Zhu, Jieming and Dong, Zhenhua and Wen, Ji-Rong},
  journal={ACM Transactions on Information Systems},
  volume={43},
  number={6},
  pages={1--47},
  year={2025},
  publisher={ACM New York, NY}
}

@inproceedings{park2023generative,
  title={Generative agents: Interactive simulacra of human behavior},
  author={Park, Joon Sung and O'Brien, Joseph and Cai, Carrie Jun and Morris, Meredith Ringel and Liang, Percy and Bernstein, Michael S},
  booktitle={Proceedings of the 36th annual acm symposium on user interface software and technology},
  pages={1--22},
  year={2023}
}

@article{gilboa2017neurobiology,
  title={Neurobiology of schemas and schema-mediated memory},
  author={Gilboa, Asaf and Marlatte, Hannah},
  journal={Trends in cognitive sciences},
  volume={21},
  number={8},
  pages={618--631},
  year={2017},
  publisher={Elsevier}
}

@article{yan2025general,
  title={General Agentic Memory Via Deep Research},
  author={Yan, BY and Li, Chaofan and Qian, Hongjin and Lu, Shuqi and Liu, Zheng},
  journal={arXiv preprint arXiv:2511.18423},
  year={2025}
}

@misc{hu2025memoryageaiagents,
      title={Memory in the Age of AI Agents}, 
      author={Yuyang Hu and Shichun Liu and Yanwei Yue and Guibin Zhang and Boyang Liu and Fangyi Zhu and Jiahang Lin and Honglin Guo and Shihan Dou and Zhiheng Xi and Senjie Jin and Jiejun Tan and Yanbin Yin and Jiongnan Liu and Zeyu Zhang and Zhongxiang Sun and Yutao Zhu and Hao Sun and Boci Peng and Zhenrong Cheng and Xuanbo Fan and Jiaxin Guo and Xinlei Yu and Zhenhong Zhou and Zewen Hu and Jiahao Huo and Junhao Wang and Yuwei Niu and Yu Wang and Zhenfei Yin and Xiaobin Hu and Yue Liao and Qiankun Li and Kun Wang and Wangchunshu Zhou and Yixin Liu and Dawei Cheng and Qi Zhang and Tao Gui and Shirui Pan and Yan Zhang and Philip Torr and Zhicheng Dou and Ji-Rong Wen and Xuanjing Huang and Yu-Gang Jiang and Shuicheng Yan},
      year={2025},
      eprint={2512.13564},
      archivePrefix={arXiv},
      primaryClass={cs.CL},
      url={https://arxiv.org/abs/2512.13564}, 
}

@misc{yao2023reactsynergizingreasoningacting,
      title={ReAct: Synergizing Reasoning and Acting in Language Models}, 
      author={Shunyu Yao and Jeffrey Zhao and Dian Yu and Nan Du and Izhak Shafran and Karthik Narasimhan and Yuan Cao},
      year={2023},
      eprint={2210.03629},
      archivePrefix={arXiv},
      primaryClass={cs.CL},
      url={https://arxiv.org/abs/2210.03629}, 
}

@misc{wang2023voyageropenendedembodiedagent,
      title={Voyager: An Open-Ended Embodied Agent with Large Language Models}, 
      author={Guanzhi Wang and Yuqi Xie and Yunfan Jiang and Ajay Mandlekar and Chaowei Xiao and Yuke Zhu and Linxi Fan and Anima Anandkumar},
      year={2023},
      eprint={2305.16291},
      archivePrefix={arXiv},
      primaryClass={cs.AI},
      url={https://arxiv.org/abs/2305.16291}, 
}

@misc{zhong2023memorybankenhancinglargelanguage,
      title={MemoryBank: Enhancing Large Language Models with Long-Term Memory}, 
      author={Wanjun Zhong and Lianghong Guo and Qiqi Gao and He Ye and Yanlin Wang},
      year={2023},
      eprint={2305.10250},
      archivePrefix={arXiv},
      primaryClass={cs.CL},
      url={https://arxiv.org/abs/2305.10250}, 
}

@misc{wang2025speculativeragenhancingretrieval,
      title={Speculative RAG: Enhancing Retrieval Augmented Generation through Drafting}, 
      author={Zilong Wang and Zifeng Wang and Long Le and Huaixiu Steven Zheng and Swaroop Mishra and Vincent Perot and Yuwei Zhang and Anush Mattapalli and Ankur Taly and Jingbo Shang and Chen-Yu Lee and Tomas Pfister},
      year={2025},
      eprint={2407.08223},
      archivePrefix={arXiv},
      primaryClass={cs.CL},
      url={https://arxiv.org/abs/2407.08223}, 
}

@article{li2023camel,
  title={Camel: Communicative agents for" mind" exploration of large language model society},
  author={Li, Guohao and Hammoud, Hasan and Itani, Hani and Khizbullin, Dmitrii and Ghanem, Bernard},
  journal={Advances in Neural Information Processing Systems},
  volume={36},
  pages={51991--52008},
  year={2023}
}

@article{zhou2023webarena,
  title={Webarena: A realistic web environment for building autonomous agents},
  author={Zhou, Shuyan and Xu, Frank F and Zhu, Hao and Zhou, Xuhui and Lo, Robert and Sridhar, Abishek and Cheng, Xianyi and Ou, Tianyue and Bisk, Yonatan and Fried, Daniel and others},
  journal={arXiv preprint arXiv:2307.13854},
  year={2023}
}

@article{wang2023query2doc,
  title={Query2doc: Query expansion with large language models},
  author={Wang, Liang and Yang, Nan and Wei, Furu},
  journal={arXiv preprint arXiv:2303.07678},
  year={2023}
}

@misc{maharana2024evaluatinglongtermconversationalmemory,
      title={Evaluating Very Long-Term Conversational Memory of LLM Agents}, 
      author={Adyasha Maharana and Dong-Ho Lee and Sergey Tulyakov and Mohit Bansal and Francesco Barbieri and Yuwei Fang},
      year={2024},
      eprint={2402.17753},
      archivePrefix={arXiv},
      primaryClass={cs.CL},
      url={https://arxiv.org/abs/2402.17753}, 
}

@article{yoran2023making,
  title={Making retrieval-augmented language models robust to irrelevant context},
  author={Yoran, Ori and Wolfson, Tomer and Ram, Ori and Berant, Jonathan},
  journal={arXiv preprint arXiv:2310.01558},
  year={2023}
}

@inproceedings{yang2018hotpotqa,
  title={HotpotQA: A dataset for diverse, explainable multi-hop question answering},
  author={Yang, Zhilin and Qi, Peng and Zhang, Saizheng and Bengio, Yoshua and Cohen, William and Salakhutdinov, Ruslan and Manning, Christopher D},
  booktitle={Proceedings of the 2018 conference on empirical methods in natural language processing},
  pages={2369--2380},
  year={2018}
}

@misc{sarthi2024raptorrecursiveabstractiveprocessing,
      title={RAPTOR: Recursive Abstractive Processing for Tree-Organized Retrieval}, 
      author={Parth Sarthi and Salman Abdullah and Aditi Tuli and Shubh Khanna and Anna Goldie and Christopher D. Manning},
      year={2024},
      eprint={2401.18059},
      archivePrefix={arXiv},
      primaryClass={cs.CL},
      url={https://arxiv.org/abs/2401.18059}, 
}

@article{guo2025deepseek,

  title={Deepseek-r1: Incentivizing reasoning capability in llms via reinforcement learning},

  author={Guo, Daya and Yang, Dejian and Zhang, Haowei and Song, Junxiao and Zhang, Ruoyu and Xu, Runxin and Zhu, Qihao and Ma, Shirong and Wang, Peiyi and Bi, Xiao and others},

  journal={arXiv preprint arXiv:2501.12948},

  year={2025}

}

@article{tsatsaronis2015overview,
  title={An overview of the BIOASQ large-scale biomedical semantic indexing and question answering competition},
  author={Tsatsaronis, George and Balikas, Georgios and Malakasiotis, Prodromos and Partalas, Ioannis and Zschunke, Matthias and Alvers, Michael R and Weissenborn, Dirk and Krithara, Anastasia and Petridis, Sergios and Polychronopoulos, Dimitris and others},
  journal={BMC bioinformatics},
  volume={16},
  number={1},
  pages={138},
  year={2015},
  publisher={Springer}
}

@article{yu2025memagent,
  title={MemAgent: Reshaping Long-Context LLM with Multi-Conv RL-based Memory Agent},
  author={Yu, Hongli and Chen, Tinghong and Feng, Jiangtao and Chen, Jiangjie and Dai, Weinan and Yu, Qiying and Zhang, Ya-Qin and Ma, Wei-Ying and Liu, Jingjing and Wang, Mingxuan and others},
  journal={arXiv preprint arXiv:2507.02259},
  year={2025}
}

@article{yang2025qwen3,
  title={Qwen3 technical report},
  author={Yang, An and Li, Anfeng and Yang, Baosong and Zhang, Beichen and Hui, Binyuan and Zheng, Bo and Yu, Bowen and Gao, Chang and Huang, Chengen and Lv, Chenxu and others},
  journal={arXiv preprint arXiv:2505.09388},
  year={2025}
}

@article{grattafiori2024llama,
  title={The llama 3 herd of models},
  author={Grattafiori, Aaron and Dubey, Abhimanyu and Jauhri, Abhinav and Pandey, Abhinav and Kadian, Abhishek and Al-Dahle, Ahmad and Letman, Aiesha and Mathur, Akhil and Schelten, Alan and Vaughan, Alex and others},
  journal={arXiv preprint arXiv:2407.21783},
  year={2024}
}

@article{kang2025memory,
  title={Memory OS of AI Agent},
  author={Kang, Jiazheng and Ji, Mingming and Zhao, Zhe and Bai, Ting},
  journal={arXiv preprint arXiv:2506.06326},
  year={2025}
}

@article{kumaran2016learning,
  title={What learning systems do intelligent agents need? Complementary learning systems theory updated},
  author={Kumaran, Dharshan and Hassabis, Demis and McClelland, James L},
  journal={Trends in cognitive sciences},
  volume={20},
  number={7},
  pages={512--534},
  year={2016},
  publisher={Elsevier}
}

@inproceedings{ma2023query,
  title={Query rewriting in retrieval-augmented large language models},
  author={Ma, Xinbei and Gong, Yeyun and He, Pengcheng and Zhao, Hai and Duan, Nan},
  booktitle={Proceedings of the 2023 Conference on Empirical Methods in Natural Language Processing},
  pages={5303--5315},
  year={2023}
}

@misc{xi2023rise,
      title={The Rise and Potential of Large Language Model Based Agents: A Survey}, 
      author={Zhiheng Xi and Wenxiang Chen and Xin Guo and Wei He and Yiwen Ding and Boyang Hong and Ming Zhang and Junzhe Wang and Senjie Jin and Enyu Zhou and Rui Zheng and Xiaoran Fan and Xiao Wang and Limao Xiong and Yuhao Zhou and Weiran Wang and Changhao Jiang and Yicheng Zou and Xiangyang Liu and Zhangyue Yin and Shihan Dou and Rongxiang Weng and Wensen Cheng and Qi Zhang and Wenjuan Qin and Yongyan Zheng and Xipeng Qiu and Xuanjing Huang and Tao Gui},
      year={2023},
      eprint={2309.07864},
      archivePrefix={arXiv},
      primaryClass={cs.AI},
      url={https://arxiv.org/abs/2309.07864}, 
}

@misc{edge2025localglobalgraphrag,
      title={From Local to Global: A Graph RAG Approach to Query-Focused Summarization}, 
      author={Darren Edge and Ha Trinh and Newman Cheng and Joshua Bradley and Alex Chao and Apurva Mody and Steven Truitt and Dasha Metropolitansky and Robert Osazuwa Ness and Jonathan Larson},
      year={2025},
      eprint={2404.16130},
      archivePrefix={arXiv},
      primaryClass={cs.CL},
      url={https://arxiv.org/abs/2404.16130}, 
}

@misc{packer2024memgptllmsoperatingsystems,
      title={MemGPT: Towards LLMs as Operating Systems}, 
      author={Charles Packer and Sarah Wooders and Kevin Lin and Vivian Fang and Shishir G. Patil and Ion Stoica and Joseph E. Gonzalez},
      year={2024},
      eprint={2310.08560},
      archivePrefix={arXiv},
      primaryClass={cs.AI},
      url={https://arxiv.org/abs/2310.08560}, 
}

@article{xi2025rise,
  title={The rise and potential of large language model based agents: A survey},
  author={Xi, Zhiheng and Chen, Wenxiang and Guo, Xin and He, Wei and Ding, Yiwen and Hong, Boyang and Zhang, Ming and Wang, Junzhe and Jin, Senjie and Zhou, Enyu and others},
  journal={Science China Information Sciences},
  volume={68},
  number={2},
  pages={121101},
  year={2025},
  publisher={Springer}
}

@article{shinn2023reflexion,
  title={Reflexion: Language agents with verbal reinforcement learning},
  author={Shinn, Noah and Cassano, Federico and Gopinath, Ashwin and Narasimhan, Karthik and Yao, Shunyu},
  journal={Advances in Neural Information Processing Systems},
  volume={36},
  pages={8634--8652},
  year={2023}
}

@article{wang2024survey,
  title={A survey on large language model based autonomous agents},
  author={Wang, Lei and Ma, Chen and Feng, Xueyang and Zhang, Zeyu and Yang, Hao and Zhang, Jingsen and Chen, Zhiyuan and Tang, Jiakai and Chen, Xu and Lin, Yankai and others},
  journal={Frontiers of Computer Science},
  volume={18},
  number={6},
  pages={186345},
  year={2024},
  publisher={Springer}
}

@article{tse2007schemas,
  title={Schemas and memory consolidation},
  author={Tse, Dorothy and Langston, Rosamund F and Kakeyama, Masaki and Bethus, Ingrid and Spooner, Patrick A and Wood, Emma R and Witter, Menno P and Morris, Richard GM},
  journal={Science},
  volume={316},
  number={5821},
  pages={76--82},
  year={2007},
  publisher={American Association for the Advancement of Science}
}

@inproceedings{shi2023large,
  title={Large language models can be easily distracted by irrelevant context},
  author={Shi, Freda and Chen, Xinyun and Misra, Kanishka and Scales, Nathan and Dohan, David and Chi, Ed H and Sch{\"a}rli, Nathanael and Zhou, Denny},
  booktitle={International Conference on Machine Learning},
  pages={31210--31227},
  year={2023},
  organization={PMLR}
}

@inproceedings{mallen2023not,
  title={When not to trust language models: Investigating effectiveness of parametric and non-parametric memories},
  author={Mallen, Alex and Asai, Akari and Zhong, Victor and Das, Rajarshi and Khashabi, Daniel and Hajishirzi, Hannaneh},
  booktitle={Proceedings of the 61st Annual Meeting of the Association for Computational Linguistics (Volume 1: Long Papers)},
  pages={9802--9822},
  year={2023}
}

@article{lu2024small,
  title={Small language models: Survey, measurements, and insights},
  author={Lu, Zhenyan and Li, Xiang and Cai, Dongqi and Yi, Rongjie and Liu, Fangming and Zhang, Xiwen and Lane, Nicholas D and Xu, Mengwei},
  journal={arXiv preprint arXiv:2409.15790},
  year={2024}
}

@inproceedings{rajpurkar2016squad,
  title={SQuAD: 100,000+ Questions for Machine Comprehension of Text},
  author={Rajpurkar, Pranav and Zhang, Jian and Lopyrev, Konstantin and Liang, Percy},
  booktitle={Proceedings of the 2016 Conference on Empirical Methods in Natural Language Processing (EMNLP)},
  year={2016}
}

@inproceedings{papineni2002bleu,
  title={BLEU: a Method for Automatic Evaluation of Machine Translation},
  author={Papineni, Kishore and Roukos, Salim and Ward, Todd and Zhu, Wei-Jing},
  booktitle={Proceedings of the 40th Annual Meeting of the Association for Computational Linguistics (ACL)},
  pages={311--318},
  year={2002}
}

@inproceedings{banerjee2005meteor,
  title={METEOR: An Automatic Metric for MT Evaluation with Improved Correlation with Human Judgments},
  author={Banerjee, Satanjeev and Lavie, Alon},
  booktitle={Proceedings of the ACL Workshop on Intrinsic and Extrinsic Evaluation Measures for Machine Translation and/or Summarization},
  year={2005}
}

@inproceedings{zhang2020bertscore,
  title={BERTScore: Evaluating Text Generation with BERT},
  author={Zhang, Tianyi and Kishore, Varsha and Wu, Felix and Weinberger, Kilian Q. and Artzi, Yoav},
  booktitle={International Conference on Learning Representations (ICLR)},
  year={2020}
}

@inproceedings{jarvelin2002cumulated,
  title={Cumulated Gain-based Evaluation of IR Techniques},
  author={J{\"a}rvelin, Kalervo and Kek{\"a}l{\"a}inen, Jaana},
  booktitle={ACM Transactions on Information Systems},
  pages={422--446},
  year={2002}
}
